\documentclass{ieeeaccess}

\usepackage{graphicx}
\usepackage[usenames,dvipsnames,svgnames,table]{xcolor}
\usepackage{caption}
\usepackage{amsmath,amssymb,enumerate}

\usepackage{amsthm}
\usepackage{mathtools}
\usepackage{breqn}
\usepackage{algorithm, algorithmicx, algpseudocode}

\usepackage{blindtext}
\usepackage{gensymb}
\usepackage{xparse}
\usepackage{lipsum}
\usepackage{mathrsfs}
\usepackage[mathscr]{euscript}
\usepackage{times}
\usepackage{cite} % to group references
\usepackage{multicol}
\usepackage[caption=false,font=normalsize]{subfig}
\usepackage{amsfonts}
\usepackage[utf8]{inputenc}
\usepackage[T1]{fontenc}
\usepackage{textcomp}
\usepackage{balance}
\usepackage{soul}
\usepackage{multirow}
\usepackage{booktabs}
\usepackage{wrapfig}
\usepackage{sidecap}
\usepackage{makecell}
\usepackage{array,float}
\usepackage[colorlinks=true,pdfpagemode=UseNone,citecolor=black,linkcolor=black,urlcolor=BrickRed]{hyperref}

\renewcommand{\thefigure}{\arabic{figure}}

\definecolor{scarColor}{rgb}{0.9608, 0.5882, 0.3922}
\definecolor{sbicycleColor}{rgb}{0.9608, 0.9020, 0.3922}
\definecolor{smotorcycleColor}{rgb}{0.5882, 0.2353, 0.1176}
\definecolor{struckColor}{rgb}{0.7059, 0.1176, 0.3137}
\definecolor{sothervehicleColor}{rgb}{1, 0.3137, 0.3922}
\definecolor{spersonColor}{rgb}{0.1176, 0.1176, 1}
\definecolor{sbicyclistColor}{rgb}{0.7843, 0.1569, 1}
\definecolor{smotorcyclistColor}{rgb}{0.3529, 0.1176, 0.5882}
\definecolor{sroadColor}{rgb}{1, 0, 1}
\definecolor{sparkingColor}{rgb}{1, 0.5882, 1}
\definecolor{ssidewalkColor}{rgb}{0.2941, 0, 0.2941}
\definecolor{sothergroundColor}{rgb}{0.2941, 0, 0.6863}
\definecolor{sbuildingColor}{rgb}{0, 0.7843, 1}
\definecolor{sfenceColor}{rgb}{0.1961, 0.4706, 1}
\definecolor{svegetationColor}{rgb}{0, 0.6863 ,0}
\definecolor{strunkColor}{rgb}{0, 0.2353, 0.5294}
\definecolor{sterrainColor}{rgb}{0.3137, 0.9412, 0.5882}
\definecolor{spoleColor}{rgb}{0.5882, 0.9412, 1}
\definecolor{strafficsignColor}{rgb}{0, 0, 1}

\usepackage{float}
\usepackage{lscape}
\graphicspath{ {./media/} }
\def\BibTeX{{\rm B\kern-.05em{\sc i\kern-.025em b}\kern-.08em
    T\kern-.1667em\lower.7ex\hbox{E}\kern-.125emX}}

\begin{document}

\history{Date of publication xxxx 00, 0000, date of current version xxxx 00, 0000.}
\doi{Pending}

% paper title
\title{Dynamic Semantic Occupancy Mapping using 3D Scene Flow and Closed-Form Bayesian Inference}

\author{\uppercase{Aishwarya Unnikrishnan}\authorrefmark{1}, \uppercase{Joey Wilson}\authorrefmark{1}, \uppercase{Lu Gan}\authorrefmark{1}, \uppercase{Andrew Capodieci}\authorrefmark{2}, \uppercase{Paramsothy~Jayakumar}\authorrefmark{3}, \uppercase{Kira Barton}\authorrefmark{1}, and \uppercase{Maani Ghaffari}\authorrefmark{1}}

\tfootnote{DISTRIBUTION A. Approved for public release; distribution unlimited. OPSEC \#5575.}
\address[1]{University of Michigan, Ann Arbor, MI 48109, USA. \texttt{\{shwarya,wilsoniv,ganlu,bartonkl,maanigj\}@umich.edu}.}% <-this % stops a space
\address[2]{Neya Systems Division, Applied Research Associates, Warrendale, PA 15086, USA. \texttt{acapodieci@neyarobotics.com}.}% <-this % stops a space
\address[3]{US Army CCDC Ground Vehicle Systems Center, Warren, MI 48397, USA. \texttt{paramsothy.jayakumar.civ@army.mil}.}

\markboth
{Aishwarya Unnikrishnan \headeretal: Dynamic Semantic Occupancy Mapping using 3D Scene Flow and Closed-Form Bayesian Inference}
{Aishwarya Unnikrishnan \headeretal: Dynamic Semantic Occupancy Mapping using 3D Scene Flow and Closed-Form Bayesian Inference}

\corresp{Corresponding author: Joey Wilson, \texttt{wilsoniv@umich.edu}.}

\begin{abstract}
This paper reports on a dynamic semantic mapping framework that incorporates 3D scene flow measurements into a closed-form Bayesian inference model. Existence of dynamic objects in the environment \textcolor{black}{can cause} artifacts and traces in current mapping algorithms, leading to an inconsistent map posterior. We leverage state-of-the-art semantic segmentation and 3D flow estimation using deep learning to provide measurements for map inference. We develop a Bayesian model that propagates the scene with flow and infers a 3D \textcolor{black}{continuous (i.e., can be queried at arbitrary resolution)} semantic occupancy map \textcolor{black}{outperforming} its static counterpart. \textcolor{black}{Extensive experiments} using publicly available data sets show that the proposed framework \textcolor{black}{improves over} its predecessors and input measurements from deep neural networks consistently.
\end{abstract}

\begin{keywords}
Bayesian Inference, Computer Vision, Mapping, Semantic Scene Understanding
% Enter key words or phrases in alphabetical 
% order, separated by commas. For a list of suggested keywords, send a blank 
% e-mail to keywords@ieee.org or visit \underline
% {http://www.ieee.org/organizations/pubs/ani\_prod/keywrd98.txt}
\end{keywords}

\titlepgskip=-15pt

\maketitle

\section{Introduction}

Mapping, localization and navigation are among the key capabilities of autonomous systems. For robots to navigate safely in complex \textcolor{black}{and evolving} environments, mapping can act as a unified framework that addresses multiple \textcolor{black}{perception} sub-tasks required for a higher-level scene understanding\textcolor{black}{, such as occupancy/traversability estimation, object detection and tracking}. While some research streams employ end-to-end deep neural networks for mapless navigation via imitation~\cite{bojarski2016end, codevilla2018end}, reinforcement~\cite{tai2017virtual, chiang2019learning} or self-supervised learning~\cite{kahn2021badgr}, maps are still widely used for explicit reliability, interpretability, and predictability. \textcolor{black}{In this work, we focus on the map inference problem instead of Simultaneous Localization and Mapping (SLAM), and aim at improving the inference performance in dynamic environments.}

Map inference can aid robots in reasoning about areas that are currently occluded but previously observed (occlusion-awareness), or inferring the geometry and semantics of an unseen landmark near those that were previously observed (smoothing). In complex environments (e.g., driving scenarios), robots can \textcolor{black}{recognize stationary cars and people, while consistently tracking} moving vehicles and pedestrians.

Semantic mapping complements geometric modelling of a robot's surroundings with semantic concepts, i.e., an understanding of what the environment \textcolor{black}{\emph{means}} to the robot. With semantic mapping, these \textcolor{black}{semantic concepts} manifest as a representation of the environment, thus lending robots more resources for task \textcolor{black}{planning and} execution. The emergence of semantic mapping can be attributed to (i) the limitations of purely geometric maps, and (ii) the advancements in deep neural networks that allow semantic interpretation of raw sensory data~\cite{garg2021semantics}. 

% Dynamic mapping is one such sub-field where the focus is on building high-level representations of the environment in the presence of mobile objects.

%\sout{However, most works construct a map explicitly and localize the robot within that map for navigation and other tasks due to the reliability, interpretability, and predictability. Furthermore, mapping also has non-substitute roles in surveillance and monitoring, scene understanding, and augmented reality.}

\begin{figure}[t]
    \centering
    \includegraphics[width=\linewidth]{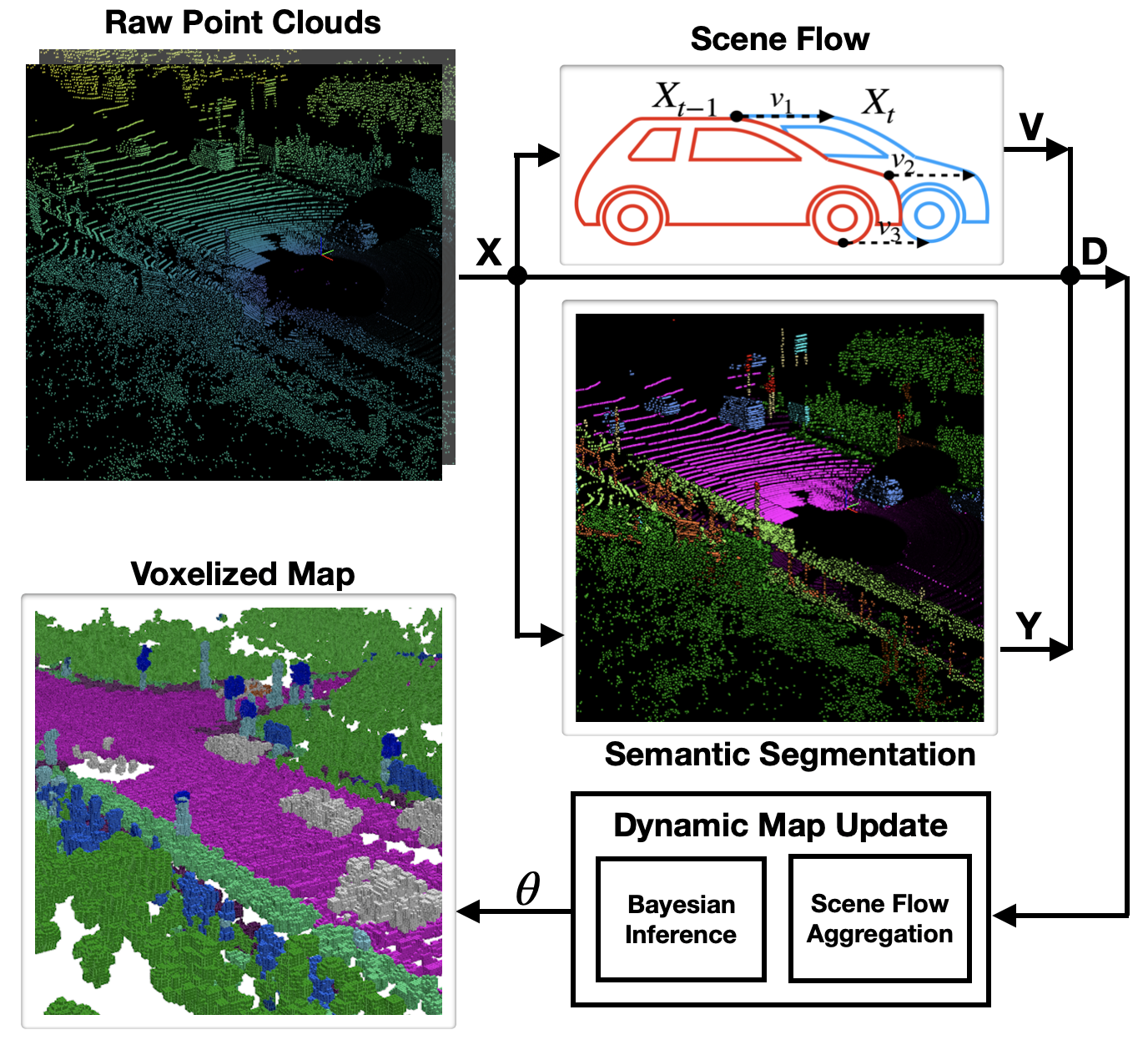}
    \caption{Dynamic Semantic Mapping Pipeline. Raw point clouds are inputs to scene flow and semantic segmentation neural networks, which compute the input to the mapping algorithm, $\mathcal{D}_t = \{\mathcal{X}_t, \mathcal{Y}_t, \mathcal{V}_t\}$. The dynamic map updates voxels parameterized by $\theta$ using scene flow aggregation and Bayesian inference. The dynamic map is capable of updating cells with dynamic objects, without leaving any residual traces.}
    \label{fig:first_fig}
\end{figure}

In \textcolor{black}{early} semantic mapping works, semantics and \textcolor{black}{geometrics are modeled independently, where semantic labels are added on top of an existing geometric representation}, such as point cloud model~\cite{sunderhauf2017meaningful}, surfel-based map~\cite{mccormac2017semanticfusion}, and voxel-based map~\cite{yang2017semantic}. As this field progresses, semantics and geometry have been modeled jointly and inferred in a unified framework~\cite{cherabier2018learning, gan2019bayesian}. Gan et al.~\cite{gan2019bayesian} proposed a unified semantic mapping framework for closed-form Bayesian inference of the semantic map posterior. However, its underlying static world assumption limits its applications in real-world \emph{dynamic} environments. In scenarios with dynamic objects, \textcolor{black}{static mapping may provide less detailed or even inconsistent reconstruction due to obscured views}. \textcolor{black}{The novelty of this work is thus to propose a unified closed-form Bayesian inference framework which extends semantic mapping to dynamic environments.}

\textcolor{black}{To this end}, we develop a scalable dynamic semantic mapping framework that combines motion and semantic information through closed-form Bayesian inference in a single pipeline as shown in Fig.~\ref{fig:first_fig}. Spatio-temporal motion data and semantic labels are aggregated over past frames and neighbouring voxels. \textcolor{black}{The aggregated motion is then used in a proposed Bayesian model to propagate the current scene and its semantic labels.}

In particular, this work has the following contributions.
% {\small
\begin{enumerate}
    \item We propose a \textcolor{black}{kernel method for scene flow aggregation and} an efficient auto-regressive Bayesian model for scene propagation.
    \item We extend the \textcolor{black}{Bayesian Kernel Inference (BKI)} semantic mapping framework~\cite{gan2019bayesian} to dynamic scenes by incorporating motion information.
    %\item The overall performance on 28 classes including dynamic objects of our method ranks the 2nd place among 69 participants in SemanticKITTI semantic segmentation multi-scan competition~\cite{behley2019semantickitti}.
    \item \textcolor{black}{We introduce an evaluation methodology for dynamic semantic mapping using single and multi-view data.}
    \item The open-source software is publicly available at: \\ {\footnotesize \href{https://github.com/UMich-CURLY/BKIDynamicSemanticMapping}{https://github.com/UMich-CURLY/BKIDynamicSemanticMapping}}
\end{enumerate}
% }

Th remaining sections are organized as follows: A comprehensive literature review is presented in Section~\ref{sec:literature}. Section~\ref{sec:preliminaries} presents the problem setup and preliminaries. The methodology is discussed in Section~\ref{sec:method}. Section~\ref{sec:eval} presents quantitative evaluation methods for dynamic mapping. Results and discussion are given in Section~\ref{sec:label}. Finally, Section~\ref{sec:conclusion} concludes the paper and \textcolor{black}{provides} ideas for future work. 

\section{Related Work}\label{sec:literature}

In this section, we review works on semantic and dynamic mapping. \textcolor{black}{While our work is focused specifically on mapping, the improved mapping algorithm can lead to improvements in downstream tasks such as localization when integrated in a SLAM system.} \textcolor{black}{Therefore, we provide background and perform comparisons of both mapping and SLAM systems to highlight the differences in how the environment is perceived, and subsequently represented.} A taxonomy of the state-of-the-art dynamic mapping works is given in Table~\ref{tab:paper_table}, based on the presence ($\checkmark$) or absence ($\times$) of semantics and scene dynamics usage, the type of sensors they operate on, and the type of flow measurements incorporated. 

\subsection{Semantic Mapping} 
Semantics are important to robot perception for better scene understanding and interaction~\cite{garg2021semantics}. \textcolor{black}{Whereas many semantic mapping works have explored learning-based local mapping \cite{LiftSplatShoot, DeepTracking, PredictingMap, MASS, PillarSegNet, MotionSC}, our work is a mathematically-derived 3D global mapping algorithm. Additionally, our method builds upon existing deep learning research by directly taking the output from neural networks as input}, \textcolor{black}{instead of attempting to embed all information within a latent space. With explicit intermediate steps, our model is diagnosable and reliable, as failures at each stage of the pipeline may be identified and observed.} For this reason, we \textcolor{black}{consider works that incorporate semantic measurements into maps given poses and estimated semantic labels}. 

Among a large body of semantic mapping studies, SemanticFusion~\cite{mccormac2017semanticfusion} can be regarded as a classic approach where the semantic probabilities of single-frame 2D images are obtained from a Convolutional Neural Network (CNN) and re-projected into 3D, after which a Bayesian update scheme fuses multi-scan probabilities \textcolor{black}{into a semantic surfel map}. Other works differ in the deep neural network used (e.g., recurrent neural networks on consecutive frames~\cite{xiang2017rnn, cheng2020robust}, 3D CNN for point clouds~\cite{dube2020segmap}), the map representation employed (point-cloud maps~\cite{sunderhauf2017meaningful, cheng2020robust} and voxel-based maps~\cite{xiang2017rnn, yang2017semantic, mccormac2018fusion++}), or the type of semantics (instance-level~\cite{grinvald2019volumetric}, object-level~\cite{sunderhauf2017meaningful, zeng2018semantic, maskfusion, detect} and place-level~\cite{sunderhauf2016place}). More recently, distributed semantic mapping for multi-robots~\cite{yue2020hierarchical, jamieson2021multi} and 3D scene graphs~\cite{rosinol20203d} are also trending research topics. 

%Added more citations to Lu's paragraph + scene graphs
% There are various map representations used to process sensor information effectively and abstract to semantic states. 
% \jw{Why surfels twice?}
% Map representations can range from re-projecting 2D static images to a point-cloud-based semantic map~\cite{cheng2020robust}, semantic oct-tree~\cite{yu2018ds,xu2019slam}, object-level representations only (semantic/non-semantic)~\cite{salas2013slam++,detect,huang2019clusterslam}, surfels~\cite{suma++}, scene graphs~\cite{rosinol2019kimera} or objects in voxels~\cite{mccormac2018fusion++}, point cloud maps or surfels~\cite{detect,maskfusion,mccormac2018fusion++}. Surfel-based representations have the limitation of not modelling occupied or free space space and therefore, cannot be used for robot navigation.

Another line of research concerns \emph{continuous semantic mapping} with uncertainty~\cite{wang2016fast, jadidi2017warped, doherty2019learning, gan2021multi}, that allows one to query maps at arbitrary resolutions. Kernel methods such as Gaussian Processes (GPs) are well-established for predicting a continuous non-parametric function to represent the semantic map~\cite{jadidi2017gaussian, zobeidi2020dense, guerrero2021sparse}. \textcolor{black}{BKI is an efficient approximation of GPs} which yields fast computation and accurate inference for semantic mapping~\cite{gan2019bayesian}. This work extends~\cite{gan2019bayesian} to dynamic scenes.

%Traversing hierarchical data structures such as oct-trees are subject to GPU thread divergence and a TSDF representation that can be accessed using a voxel hashing scheme~\cite{neissner2013} is adapted \cite{vespa2018efficient,laine2010efficient} to semantic SLAM methods~\cite{xu2019mid,strecke2019fusion,barsan2018robust}. Our work builds upon a voxel map, within which an oct-tree is rooted. The nodes of the oct-tree and the voxels in 3D space can be accessed via voxel hashing.

\begin{table}[t]
    \centering
    % \resizebox{\linewidth}{!}{
    \footnotesize
    \begin{tabular}{|l|p{0.85cm}|m{1.2cm}|m{1.25cm}|m{0.6cm}|}
    \hline
    \centering
       Paper & Semantic & Scene Dynamics Retention & Sensor & Flow \\ \hline
        \multicolumn{5}{|c|}{SLAM} \\
       \hline
       DynaSLAM \cite{dynaslam} & $\times$ & $\times$ & C & $\times$ \\ % 2018
       Alcantarilla et. al. \cite{alcantarilla2012combining} & $\times$ & $\times$ & stereo & Scene \\ % 2012
       DSOD \cite{ma2019dsod} & $\times$ & $\times$ & mono & $\times$ \\ % 2019
       SOF-SLAM \cite{cui2019sof} & $\checkmark$ & $\times$ & RGB-D & Optical \\ % 2019
       DS-SLAM \cite{yu2018ds} & $\checkmark$ & $\times$ & RGB-D & Optical \\
       Brasch. et. al. \cite{brasch2018semantic} & $\checkmark$ & $\times$ & mono & $\times$ \\
       SLAM++ \cite{salas2013slam++} & $\times$ & $\times$ & RGB-D & $\times$ \\
       
       DOS-SLAM \cite{xu2019slam} & $\checkmark$ & $\times$ & RGB-D & Scene \\
       
       Detect-SLAM \cite{detect} & $\checkmark$ & $\checkmark$ & RGB-D & $\times$ \\ %propagated moving probability
       SLAMANTIC \cite{schorghuber2019slamantic} & $\checkmark$ & $\checkmark$ & mono,stereo & $\times$\\
       
       Fusion++ \cite{mccormac2018fusion++} & $\checkmark$ & $\checkmark$ & RGB-D & $\times$ \\
       MaskFusion \cite{maskfusion} & $\checkmark$ & $\checkmark$ & RGB-D & $\times$ \\
       MID-Fusion \cite{xu2019mid} & $\checkmark$ & $\checkmark$ & RGB-D & $\times$ \\
       ClusterSLAM \cite{huang2019clusterslam} & $\times$ & $\checkmark$ & stereo & $\times$ \\
       DynSLAM \cite{barsan2018robust} & $\times$ & $\checkmark$ & stereo & Scene \\
       EM-Fusion \cite{strecke2019fusion} & $\checkmark$ & $\checkmark$ & RGB-D & $\times$\\
       Rosinol et. al. \cite{rosinol20203d} & $\checkmark$ & $\checkmark$ & stereo & $\times$ \\
       Vespa et. al. \cite{vespa2018efficient} & $\times$ & $\checkmark$ & RGB-D &  $\times$\\
       Henein et. al. \cite{henein2020dynamicslam} & $\times$ & $\checkmark$ & RGB-D & $\times$  \\
       \hline
       \multicolumn{5}{|c|}{Mapping}\\\hline
       Sun et. al. \cite{sun2018recurrent} & $\checkmark$ & $\checkmark$ & LiDAR & $\times$ \\ %dynamic objects remain in the map for 5 minutes during mapping (how do they detect whether the object is dynamic? with objectness detection + they have a napping time between between observations + non-observations) % learning on different timescales
       Kochanov et. al. \cite{kochanov2016scene} & $\checkmark$ & $\checkmark$ & stereo & Scene\\ %posterior sampling from each voxel with scene flow + one-to-one correspondence check individually, update of belief, clearing belief on free cells on the basis 
       Suma++ \cite{suma++} & $\checkmark$ & $\times$ & LiDAR & $\times$ \\
       \textbf{Ours} & $\checkmark$ & $\checkmark$ & LiDAR,3DC  & Scene\\ \hline
    \end{tabular}
    % }
   \caption{\textcolor{black}{Comparison of properties of DynamicSemanticBKI with respect to other dynamic SLAM and dynamic mapping systems. Although we compare D-BKI only to other mapping baselines, we elaborate on properties of both SLAM and mapping systems to highlight the dynamic mapping taxonomy.}  In the table, C = (mono, stereo, RGB-D) and 3DC = (stereo, RGB-D).}
   
   %We describe whether the research cited (i) builds semantic maps, (ii) retains scene dynamics information in the map, (iii) which sensor each work reports, and (iv) which kind of flow estimation is used to aid dynamic mapping. Our work is a dynamic-semantic mapping framework that uses scene flow data and can operate on any sensor one can generate point-clouds from such as LiDAR, stereo and RGB-D. 
    \label{tab:paper_table}
\end{table}

\subsection{Mapping in Dynamic Environments}
Dynamic objects can break the assumption of scene rigidity in most \textcolor{black}{mapping algorithms} and cause failure. \textcolor{black}{When combined with localization in a SLAM pipeline, artifacts left by dynamic objects can introduce errors for downstream tasks such as pose estimation and loop closure.} Thus, some \textcolor{black}{SLAM} systems treat dynamic objects in a scene as spurious data or outliers, excluding them entirely from pose estimation and mapping to achieve better accuracy and robustness~\cite{dynaslam, alcantarilla2012combining, ma2019dsod}. \textcolor{black}{However, discarding dynamic objects ultimately relies upon the ability to reject dynamic objects and decreases the level of scene understanding embedded within the map. In this section, we provide background on rejection-based approaches as well as mapping algorithms which jointly model the static world and dynamic objects.}

Discarding dynamic objects may be performed through probabilistic outlier rejection~\cite{brasch2018semantic}, moving consistency check~\cite{yu2018ds}, feature-based filtering~\cite{dynaslam}, \textcolor{black}{measurement-map} semantic inconsistency check~\cite{suma++}, culling out with object-camera relative poses~\cite{salas2013slam++}, semantic and geometric information coupling~\cite{cui2019sof}, or residual motion likelihood calculation~\cite{alcantarilla2012combining, xu2019slam}. \textcolor{black}{These methods can partially reduce localization error brought by dynamic objects, but still have limitations. For instance,} discarding information based on semantic labels completely depends on the prediction accuracy. Moreover, \textcolor{black}{the discarded motion information, if modeled correctly, could be further leveraged to predict the scene dynamics.}

\textcolor{black}{There are two primary approaches to integrating motion within maps.} In the first method, scene dynamics are incorporated into a single reconstruction volume. This could be done by maintaining an object point cloud with a moving probability~\cite{detect}, calculating a dynamics factor for classes that could be mis-classified as ``dynamic'' (e.g., parked cars) and incorporating those into pose estimation~\cite{schorghuber2019slamantic}, propagating feature points by sampling from scene flow measurements~\cite{kochanov2016scene}, or fusing semantic features by recurrent observation average pooling in an OctoMap cell~\cite{sun2018recurrent}. Instead of analyzing the motion properties of map cell or feature from a single scan, we combine spatio-temporal motion data over multiple scans and neighbouring voxels.

The second category \textcolor{black}{is characterized by its underlying object-oriented map representation}. These approaches track local objects using \textcolor{black}{Iterative Closest Point (ICP)} and semantic segmentation-aided fusion~\cite{mccormac2018fusion++, maskfusion, xu2019mid}, or by clustering on the basis of motion estimation~\cite{huang2019clusterslam}. Object tracking is also done via sparse scene flow estimation~\cite{barsan2018robust}, frame-to-model data association~\cite{rosinol2019kimera} or \textcolor{black}{Signed Distance Function} (SDF)-based data association~\cite{strecke2019fusion}. In this area, deep learning-based instance segmentation is often the bottleneck for computational efficiency~\cite{dynaslam, yu2018ds, xu2019slam, detect}. 
{Although we model the scene as a unified reconstruction volume as in the former approaches, we can incorporate flow during environmental perception as in the latter approaches.}
\section{Semantic Bayesian Kernel Inference}
\label{sec:preliminaries}

\textcolor{black}{Semantic-Bayesian Kernel Inference (BKI)~\cite{gan2019bayesian} is a probabilistic method for 3D semantic mapping with quantifiable uncertainty.}The Semantic-BKI framework \textcolor{black}{assumes that the $j$-th map cell (voxel in 3D, indexed by \mbox{$j \in \mathbb{Z}^+$})} with semantic probability \mbox{${\theta}_j = (\theta_j^1, ..., \theta_j^K)$}, where $\theta_j^k$ is the probability of the $j$-th cell belonging to the $k$-th category, has the Categorical likelihood \mbox{$p(y_i \mid {\theta}_j ) = \prod_{k=1}^K (\theta_j^k)^{[y_i = k]}$}. Here,
% \begin{equation}
% \label{eq:categorical_likelihood}
%     p(y_i | {\theta}_j ) = \prod_{k=1}^K (\theta_j^k)^{[y_i = k]},
% \end{equation}
 \mbox{$y_i \in \{1, ..., K\}$} is the semantic measurement at position \mbox{$x_i \in \mathbb{R}^3$} in or around the $j$-th cell, and \mbox{$[y_i = k]$} evaluates to 1 if \mbox{$y_i = k$}, 0 otherwise. The semantic measurement $y_i$ is usually the semantic label output by a neural network. Given training data with $N$ measurement points \mbox{$\mathcal{D} = \{(x_i, y_i)\}_{i=1}^N$}, semantic mapping seeks the \emph{posterior} distribution \mbox{$p({\theta}_j \mid \mathcal{D})$} for each map cell~$j$.

% The centre of each map cell is represented by $X_j \in \mathbb{R}^3$. A parameter $\theta_j^k$ native to that map cell encodes the probability of belonging to a semantic class k out of K distinct classes. Incoming training data $Z = (x_i, y_i)_{i=1}^N$, where $x_i \in \mathbb{R}^3$ is the position at which a training point is observed and $y_i$ the corresponding semantic category, is incorporated into a map cell $X_j$ with the following categorical distribution:
% \begin{equation}
%     P(y_i | \theta_j ) = \prod_{k=1}^K (\theta_j^k)^{y_i = k}  
% \end{equation}

For a closed-form solution, BKI semantic mapping adopts a conjugate \emph{prior} over ${\theta}_j$, given by a Dirichlet distribution $\textnormal{Dir}(K, {\alpha}_0)$, \mbox{${\alpha}_0 = (\alpha_0^1, ..., \alpha_0^K)$}, where \mbox{$K \geq 2$} is the number of categories, and \mbox{$\alpha_0^k \in \mathbb{R}^+$} are concentration parameters for each category. Applying Bayes' rule and Bayesian kernel inference, the posterior is another Dirichlet distribution, given by \mbox{$\textnormal{Dir}(K, {\alpha}_j)$}, \mbox{${\alpha}_j = (\alpha_j^1, ..., \alpha_j^K)$}, and:
\begin{equation}
\label{eq:spatialupdate}
    \alpha_j^k = \alpha_0^k + \sum_{i=1}^N \mathcal{K}_s(x_i, x_j) [y_i = k],
\end{equation}
where \mbox{$\mathcal{K}_s: \mathbb{R}^3 \times \mathbb{R}^3 \rightarrow [0, 1]$} is a \emph{spatial} kernel function defined on 3D Euclidean space to capture the spatial correlation of two 3D positions, and $\alpha_j^k$ is the $k$-th concentration parameter of the \emph{query} voxel $j$ centered at $x_j \in \mathbb{R}^3$.%\mgj{Why $x_i$ is small letter and $X_j$ is capital? It's better to be consistent and use lower case (preferably) or uppercase for all 3D points.}

Given ${\alpha}_j$, the maximum a posteriori (MAP) estimate of ${\theta}_j$ can then be computed in closed-form:
\begin{equation}
\label{eq:mode}
    \hat{\theta}_j^k = \frac{\alpha_j^k - 1}{\sum_{k=1}^K \alpha_j^k - K}, \quad \alpha_j^k > 1.
\end{equation}
% The expected value and variance of $\theta_j^k$ are given as 
% \begin{equation}
% \label{eq:mean-variance}
%     \mathbb{E}[\theta_j^k]
%     = \frac{\alpha_j^k}{\sum_{k=1}^K \alpha_j^k}, \quad \mathbb{V}[\theta_j^k] = \frac{\frac{\alpha_j^k}{\sum_{k=1}^K \alpha_j^k}  (1 - \frac{\alpha_j^k}{\sum_{k=1}^K \alpha_j^k}) }{\sum_{k=1}^K \alpha_j^k + 1}.
% \end{equation}
% \begin{equation}
% \label{eq:varupdate}
%     \mathbb{V}[\theta_\ast^k] = \frac{\frac{\alpha_\ast^k}{\sum_{k=1}^K \alpha_\ast^k}  (1 - \frac{\alpha_\ast^k}{\sum_{k=1}^K \alpha_\ast^k}) }{\sum_{k=1}^K \alpha_\ast^k + 1}.
% \end{equation}
% where the variance is inversely proportional to the total number of observations.

In BKI semantic mapping, the prior distribution of the map at time stamp $t$ is directly set to be the posterior at time stamp \mbox{$t - 1$} (assuming that the \textcolor{black}{environment} does not change between two time stamps), i.e., \mbox{$p({\theta}_{j, t} \mid \mathcal{D}_{1:t-1}) = p({\theta}_{j, t-1} \mid \mathcal{D}_{1:t-1})$}, to allow recursive Bayesian updates using sequential training data:
\begin{align}
\label{eq:static_assumption}
    p( \theta_{j,t} \mid \mathcal{D}_{1:t}) &\propto p(\mathcal{D}_t \mid {\theta}_{j, t}, \mathcal{D}_{1:t-1}) p({\theta}_{j, t} \mid \mathcal{D}_{1:t-1}) \nonumber \\
    & = p(\mathcal{D}_t \mid {\theta}_{j,t}) p({\theta}_{j, t-1} \mid \mathcal{D}_{1:t-1}).
\end{align}
However, this assumption is easily violated by moving objects in the environment or environmental changes. As such, we model the transition from \mbox{$p({\theta}_{j, t-1} \mid \mathcal{D}_{1:t-1})$} to \mbox{$p({\theta}_{j, t} \mid \mathcal{D}_{1:t-1})$} using spatial and temporal information.

% This equation captures the likelihood function for the semantic category a map cell belongs to. We seek to calculate the \textbf{posterior distribution $P( \theta_j | D)$} to find a distribution over the parameter $\theta_j$ for each map cell $X_j$. To calculate the posterior distribution with Bayes' Rule in closed form, we take a Dirichlet distribution D(K, $\alpha$) as the conjugate prior. $\alpha$ is the vector that contains a hyperparameter associated with each semantic category. 

% To estimate the parameters $\theta_*$ of a query point $X_*$, we can use the extended likelihood function to relate $X_*$ to a set of observations $Z_i= (x_i, y_i)$ (the scene flow is purposely removed here). The hyperparameter $\alpha_*^k$ of the posterior distribution is estimated as:
%  \begin{equation}
%      \alpha_*^k = \alpha_{0}^k + \sum_i^N \mathcal{K}_s(x_i, X_*) y_i \label{eq:spatialupdate}
% \end{equation} 
% where $\mathcal{K}_s(x_i, X_*)$ is a spatial kernel relating query point $X_*$ to $x_i$ and $\alpha_{0}^k$ is the prior.
% In practice, the map representation is in the form of voxels, inside which there is an octree of fixed depth. The centres of the octree act as query points in Equation \ref{eq:spatialupdate}.

% \prompt{A good visualization of the octree and query point will be helpful for readers}
\section{Method: Dynamic-BKI}
\label{sec:method}
% \prompt{Motivation for section}
% An assumption made in Equation \ref{eq:spatialupdate} is that training data is independently and identically distributed over time for query point $X_\ast$. This assumption is challenged when there are dynamic objects moving around a query point. The maximum a posteriori estimate of a category k in Equation \ref{eq:mode} depends on the number of observations encountered for that semantic category and the number of observations encountered in total. As the number of observations, $\alpha_\ast^k$, contributed toward a semantic category k increases while no other observations from other categories are seen, the variance in Equation \ref{eq:varupdate} decreases. As a result, when observations $\alpha_\ast^q$ from a new semantic category q are encountered, it becomes harder to incorporate these new observations, when the confidence in category k is very high. Therefore, an alternate approach needs to be taken to distinguish between static and dynamic training data. \lu{I think this is better in Introduction.}
In this section, we introduce a method to extend Semantic-BKI to dynamic environments. We first formulate an auto-regressive temporal transition model which propagates the map posterior according to the scene dynamics. Next, we show how we aggregate motion information from the training data for incorporation into the map voxels. Finally, we consolidate and summarize the algorithm for dynamic semantic mapping.

\subsection{Temporal Transition Model}
\label{sec:transition}

\begin{figure}[t]
    \centering
    \includegraphics[width=\linewidth]{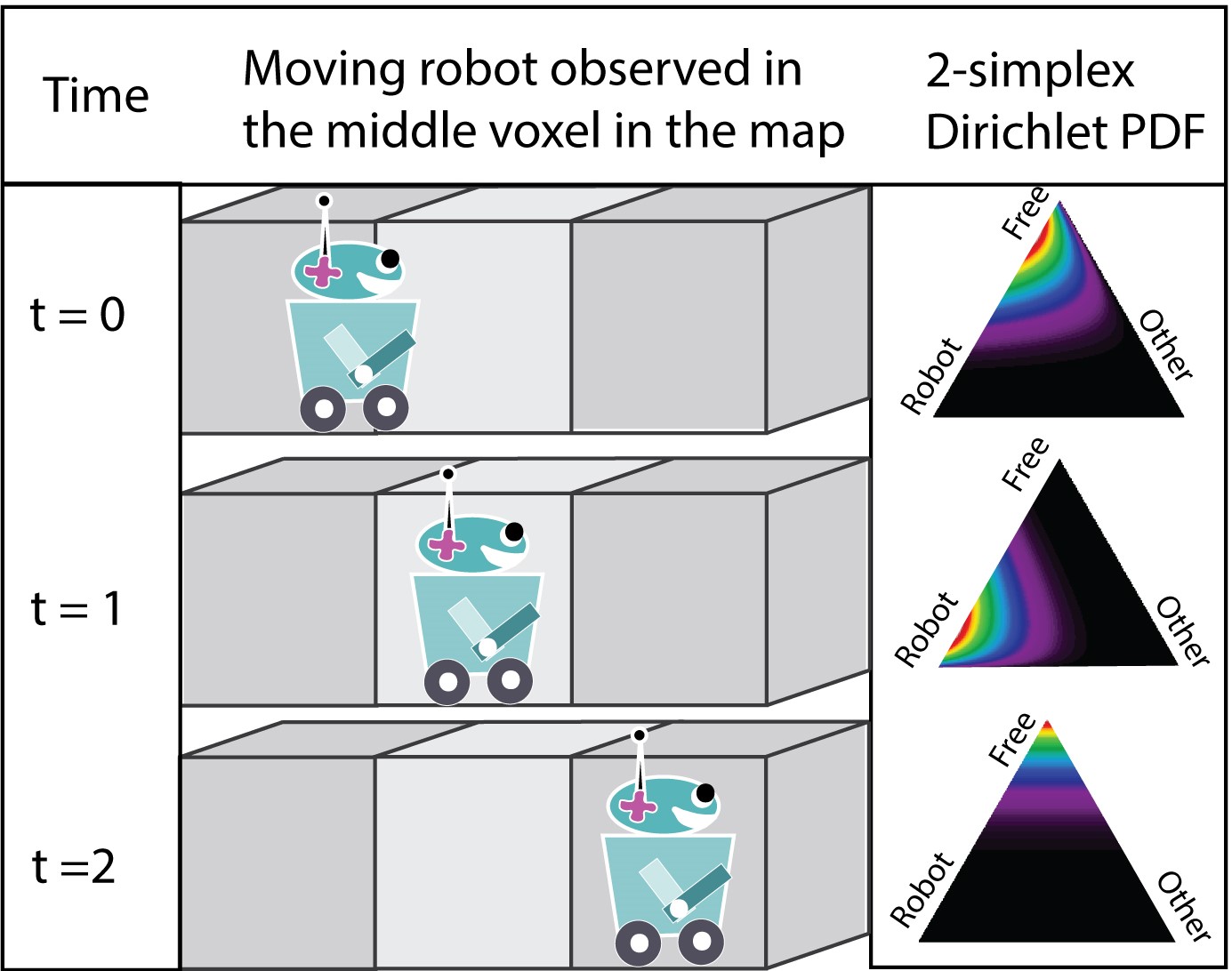}
    \caption{We illustrate the observation of a moving object through the middle voxel in this map and display how that voxel's semantics are different at every time step. For every time step, we plot the posterior Dirichlet probability density function (PDF) of the voxel on a 2-simplex. The shift in the rainbow gradient demonstrates what the belief about a semantic category (robot, free or other) can be over t=0 to t=2 to classify the voxel correctly at that time. This shift can be influenced by changing the concentration parameters (hyperparameters) of the Dirichlet posterior as: ${\alpha}_{j,0} \rightarrow {\alpha}_{j,1} \rightarrow {\alpha}_{j,2} $.}
    \label{fig:methodmotivation}
\end{figure}

When dynamic objects move in and out of a voxel $j$, the samples observed in it across time are not \textcolor{black}{independently and identically distributed \emph{(i.i.d.)}}. Samples drawn from the map posterior at different time stamps will come from independent but not identically distributed Dirichlet distributions. In Figure~\ref{fig:methodmotivation}, we illustrate the motion of an object and a corresponding visualization of the Dirichlet probability density function (PDF) over the 2-simplex when there are just three classes --- ``robot'', ``free space'', and ``other''. The static world assumption in \eqref{eq:static_assumption} solely relies on the frequency of observations in a voxel. This property makes the Dirichlet distribution ignore scene dynamics and become overconfident about classes that contribute more observations over \emph{all} time stamps rather than the \emph{current} time stamp. Therefore, for correct classification, the hyperparameters of the Dirichlet distribution under a static world assumption have to evolve with the scene dynamics.

\textcolor{black}{We first introduce notations used in our formulation.} Let the set of all classes be \textcolor{black}{$\mathcal{P}$}, the set of moving classes \textcolor{black}{be $\mathcal{Q}$} \textcolor{black}{($q \in \mathcal{Q}$)} and the free voxel category be denoted as ``free.'' Additionally, let the set of all classes \textcolor{black}{excluding a class $r$} \textcolor{black}{be $\mathcal{P} \setminus r$}. \textcolor{black}{We define a voxel flow vector for each voxel as $ v_j = (v_j^1, \ldots, v_j^K)$, which is the motion information captured per semantic category within a voxel $j$. Details on computing voxel flow is presented in Section~\ref{sec:3dflow}.} %\mgj{dynamic? otherwise it should be $\mathcal{L}\backslash q$.} 

\begin{SCfigure}
    \centering
    \includegraphics[width=0.45\linewidth]{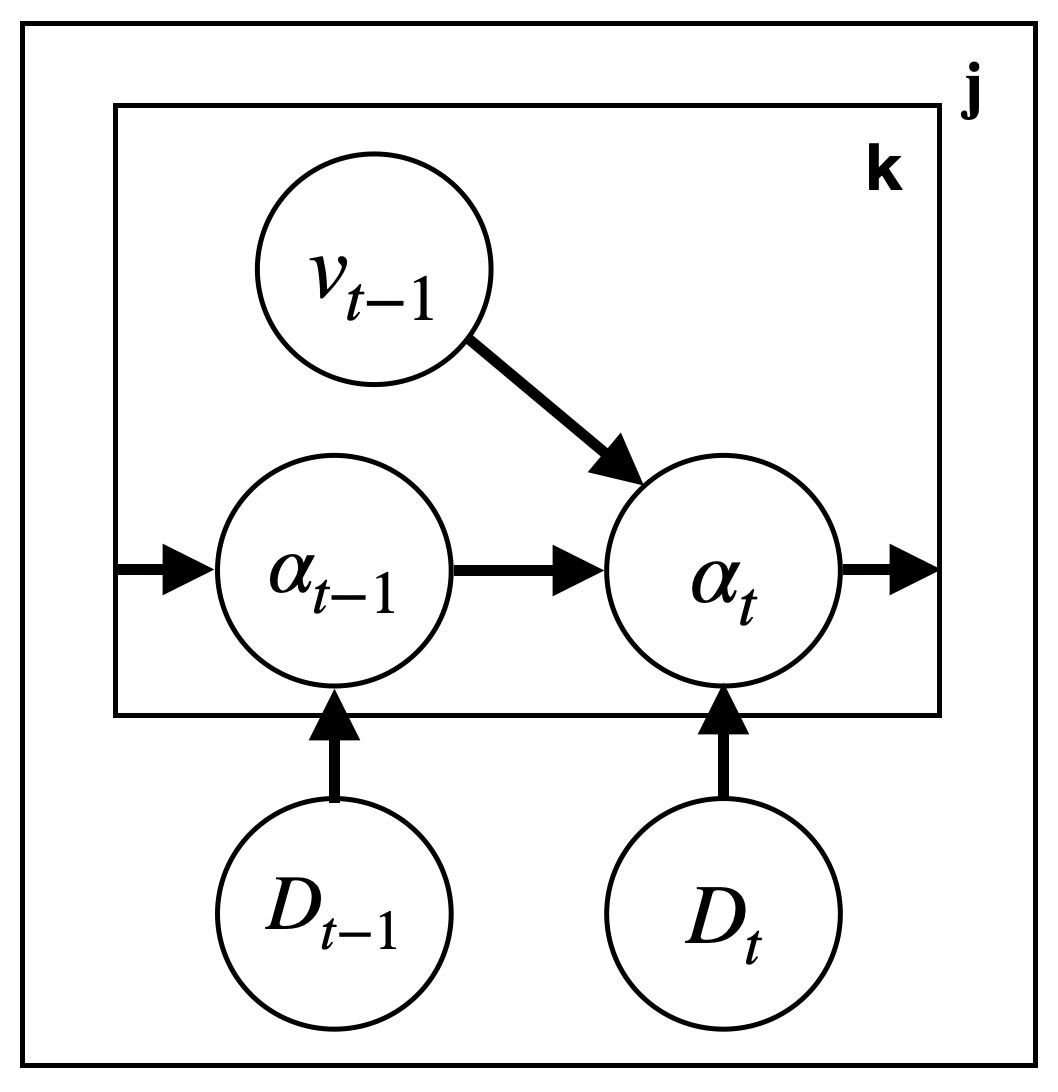}
    \caption{A graphical model for hyperparameter propagation. For each voxel $j$ updated at time $t$ and for each class $k$, the hyperparameter $\alpha_t$ is a deterministic function of flow in the previous observed time stamp $v_{t-1}$ and the prior $\alpha_{t-1}$.}
    \label{fig:temporal_model}
\end{SCfigure}

We propose a time-series model to account for temporal discrepancies in the Dirichlet distribution \textcolor{black}{caused by moving objects}. To forecast $\overline{{\alpha}}_{j,t}$ of voxel~$j$ when a moving object passes through at time stamp $t-1$, we apply an auto-regressive (AR) model that leverages the 3D motion information captured from the environment and applies it to the map prior. The \textcolor{black}{class-wise} AR model is as follows:
\begin{equation}
\overline{\alpha}_{j,t}^k = e^{-(v_{j,t-1}^k)^2} \alpha_{j,t-1}^k, \label{eq:trans_model}
\end{equation}
where $e^{-(v_{j,t-1}^k)^2}$ is the AR model's parameter, $\alpha_{j,t-1}^k$ is prior concentration parameter for class $k$ and $v_{j,t-1}^k$ is the \textcolor{black}{voxel flow at time stamp $t-1$} which influences the hyperparameter $\alpha_{j,t}^k$ for semantic class $k$.
We depict a graphical model for this temporal transition model in Figure~\ref{fig:temporal_model}.

The concept behind the transition model is to redistribute the probability mass of the concentration parameters when there is motion observed in the environment. Therefore, we warp the concentration parameters according to the effect that the motion of a dynamic object has on \emph{1)} its corresponding class and \emph{2)} other classes. Keeping these two \textcolor{black}{factors} in mind, we introduce two modules to predict the concentration parameters $\overline{{\alpha}}_{j,t}$ for voxel $j$ at time stamp $t$.

% \noindent \textbf{\emph{Backward (exit) correction [BACC]}}: 
\subsection{Backward or exit correction (BACC)}
\textcolor{black}{When a moving object of category $q$ is detected in voxel $j$ at time stamp $t-1$ is in motion and could \emph{exit} in time stamp $t$, we want to \emph{decay} its influence on the concentration parameters in $j$ for category $q$ in the upcoming $t$. As a result, we reduce the influence of prior parameter $\alpha_{j,t-1}^q$ of $q$ on concentration parameter of the next time stamp $\overline{\alpha}_{j, t}^q$. To do so, we only need to calculate the voxel flow associated with that moving category $q$, i.e., $v_{j,t-1}^q$.} The map prior (observations) of other classes \textcolor{black}{$\alpha_{j,t-1}^{k \in (\mathcal{P} \setminus q)}$} is not required as each semantic category is updated independently.

% \noindent \textbf{\emph{Forward (entry) correction [FORC]}}: 
\subsection{Forward or entry correction (FORC)}\label{sec:forc}
\textcolor{black}{When a voxel $j$ that was ``free'' in time stamp $t-1$ has a moving object of category $q \in \mathcal{Q}$ \emph{entering} it in time stamp $t$, we want to make sure that the future presence of the object can be represented in $j$. As a result, we need to reduce the effect of $\alpha_{j,t-1}^{\text{free}}$ on $\overline{\alpha}_{j,t}^{\text{free}}$ and for this, we would need to calculate voxel flow associated with the category ``free'', i.e., $v_{j,t-1}^{\text{free}}$. Intuitively, we do this with the motion information of \emph{all moving objects} in the vicinity of voxel $j$ that could enter voxel $j$. Additionally, we could use the same voxel flow $v_{j,t-1}^{\text{free}}$ to decay concentration parameters of all the static classes $\mathcal{P} \setminus \mathcal{Q}$ in order to make $\alpha_{j,t}^q$ the highest after entry.} 

\subsection{Voxel Flow Calculation from Point Clouds}\label{sec:3dflow}

Given a voxel $j$ \textcolor{black}{centered at $x_j$}, we wish to get a low-level understanding of how an object is moving in or out of it to model $v_{j,t-1}^k$ in~\eqref{eq:trans_model}. Scene flow provides us with the underlying 3D motion field of the points in the scene. Given two incoming point clouds $\mathcal{X}_{t-1}$ and $\mathcal{X}_t$, recorded at time stamp $t-1$ and $t$, respectively, we require a translational motion vector $u_i \in \mathbb{R}^3$ that conveys how much a point $x_i \in \mathcal{X}_{t-1}$ has displaced to a new location $x'_i \in \mathcal{X}_t$. \textcolor{black}{In practice, this translational motion vector can be obtained from the ``scene flow'' associated with each point in the point cloud \cite{flownet3d, battrawy2020deeplidarflow, hplflownet}.} 

To capture the \textcolor{black}{voxel flow} $ v_j = (v_j^1, \ldots, v_j^K)$ pertaining to any semantic category \textcolor{black}{for voxel $j$}, we aggregate the flow from training points around \textcolor{black}{voxel centroid $x_j$}. Thus, given training points \mbox{$\mathcal{D} = \{(x_i, y_i,  u_i)\}_{i=1}^N$}, a \textcolor{black}{spatial} kernel \mbox{$\mathcal{K}_v: \mathbb{R}^3 \times \mathbb{R}^3 \rightarrow[0, 1]$} is used to weight the influence of each point $x_i$ on \textcolor{black}{$v_j^q$} so that the closer a dynamic object of class $q \in \mathcal{Q}$ is to the voxel center, the more influence it has. \textcolor{black}{Mathematically, }this becomes a kernel density estimation problem and the per-class \textcolor{black}{voxel} flow is calculated as:
\begin{equation}
    \textcolor{black}{v_j^q} =\frac{1}{N} \sum_{i=1}^N \mathcal{K}_v(x_i, x_j) \textcolor{black}{\lVert u_i \rVert}^{[y_i = \textcolor{black}{q}]}, \label{eq:velocity}
\end{equation}
\textcolor{black}{where $\lVert \cdot \rVert$ takes the Euclidean norm. We take the vector norm consistent with $v_j^q$'s usage in the exponential AR-model in \eqref{eq:trans_model}. Therefore, our objective is to get a quantitative estimate of how much motion there is around the voxel, rather than capture the direction of motion.}
%[I remembered you have a square here??] @Lu it was always just the norm. Just checked the previous manuscript.}
\begin{algorithm}[t]
\caption{Scene Flow Aggregation}
\label{al:sceneflow}
\small 
\begin{algorithmic}[1]

\State \textbf{Input:} Training data $\mathcal{D}_t: (\mathcal{X}_t, \mathcal{Y}_t, \textcolor{black}{\mathcal{U}_t)}$ with N points; \textbf{Query point}: $x_j$; \textbf{Previous \textcolor{black}{voxel} flow estimate} ${v}_{j,t-1}$ 
% \Statex
\Procedure{AggregateFlow}{$\mathcal{D}_t, x_j, {v}_{j,t-1}$}
%\Procedure{Scene Flow Aggregation}{}\label{alg:proc1}
\For{each $(x_i, y_i, \textcolor{black}{u_i}) \in (\mathcal{X}_{t}, \mathcal{Y}_{t}, \textcolor{black}{\mathcal{U}_{t}})$} 

    \State $w_v \gets \mathcal{K}_v(x_i, x_j)$ \Comment{$x_i$'s influence on BACC}\label{alg:vback}
    \State $w_v^{\text{free}} \gets \mathcal{K}_v^{\text{free}}(x_i, x_j)$ \Comment{$x_i$'s influence on FORC}\label{alg:vfor}
    \For{\textcolor{black}{$q \in \mathcal{Q}$}} \Comment{for all dynamic classes}
        \State $v_{j, t}^q \gets  v_{j, t}^q +  w_v \textcolor{black}{\lVert u_i \rVert}^{[y_i=q]}$ \Comment{apply \eqref{eq:velocity}}
        \State $v_{j, t}^{\text{free}} \gets v_{j, t}^{\text{free}} + w_v^{\text{free}} \textcolor{black}{\lVert u_i \rVert}^{[y_i=q]}$ \Comment{apply \eqref{eq:velfree}}
    \EndFor
\EndFor
\For{\textcolor{black}{$q \in \mathcal{Q}$}}
    \State $v_{j, t}^q \gets f(\frac{v_{j, t}^q}{N}, v_{j, t-1}^q)$ \label{alg:filter} \Comment{Apply a custom filter \textcolor{black}{$f(.,.)$ with} past flow }
\EndFor
\State $v_{j, t}^{\text{free}} \gets f(\frac{v_{j, t}^{\text{free}}}{N}, v_{j, t-1}^{\text{free}})$\label{alg:filter2} %\Comment{Normalize with N, weighted average $v_{j,t-1}$ }
\For{\textcolor{black}{$k \in \mathcal{P} \setminus \mathcal{Q}$}} \Comment{\textcolor{black}{copy over} free velocity to static classes}
\State $v_{j, t}^k \gets v_{j,t}^{\text{free}}$ 
\EndFor

%potential half-update
\State \textbf{return} ${v}_{j, t}$
\EndProcedure
\end{algorithmic}
\end{algorithm}

In Section \ref{sec:forc}, we introduced FORC for free and other static classes. \textcolor{black}{As explained previously, we calculate their per-class voxel flow together by considering the dynamic objects of \emph{all} categories moving in voxel $j$}:
\begin{equation}
        v_j^{\text{free}} =\frac{1}{N} \sum_{i=1}^N \mathcal{K}_v^{\text{free}}(x_i, x_j) \textcolor{black}{\lVert u_i \rVert}^{[y_i \in \textcolor{black}{\mathcal{Q}}]}, \label{eq:velfree}
\end{equation}
%\prompt{Can you add more info on the velocity kernel? Also, is the kernel different for free vs. dynamic classes?}
where \textcolor{black}{$\mathcal{Q}$} is the set of dynamic classes and $\mathcal{K}_v^{\text{free}}$ is a \textcolor{black}{special case of $\mathcal{K}_v$} that weights the influence of dynamic training point $x_i$ on \textcolor{black}{$v_j^{\text{free}}$}. Specific details about $\mathcal{K}_v$ and $\mathcal{K}_v^{\text{free}}$ will be discussed in Section~\ref{sec:experiments} %\textcolor{red}{Check! there is a 1. }.

In Algorithm \ref{al:sceneflow}, we \textcolor{black}{summarize} how the per-class \textcolor{black}{voxel} flow for a query \textcolor{black}{voxel $j$} is estimated using the positional $\mathcal{X}_t$, semantic $\mathcal{Y}_t$, and egomotion-compensated  $\mathcal{U}_t$ information of each point \textcolor{black}{$x_i$} in a point cloud. For BACC, we aggregate the flows of the training points encountered around $x_j$ in line \algref{al:sceneflow}{alg:vback}. For FORC, we aggregate the flows of the training points while weighing the ones in neighbouring voxels more (than in BACC) in line \algref{al:sceneflow}{alg:vfor}. Whereas both equations have a similar form, BACC and FORC have separate kernels. Additionally, while FORC considers all neighboring dynamic points when updating $v_{j, t}^{\text{free}}$, BACC only computes $v_{j, t}^q$ from dynamic points with matching semantic label $q$. After calculating, $v_{j,t}^k$ for any class $k$, in lines \algref{al:sceneflow}{alg:filter} and \algref{al:sceneflow}{alg:filter2}, we post-process $v_{j,t}^k$ with a filter $f : \mathbb{R} \times \mathbb{R} \rightarrow{\mathbb{R}} $ to aggregate information from the voxel flow in the previous time step $v_{j,t-1}^q$ in the final calculation of $v_{j,t}^q$. %\textcolor{red}{[s doesn't appear to be specified anywhere. Also it may break equation (5).]}

\subsection{Map Posterior Update for Scene Propagation}\label{sec:consolidate}

Section \ref{sec:transition} describes how we account for the change in concentration parameters of the Dirichlet distribution caused by the motion of objects. \textcolor{black}{Using this model and following} a Bayesian approach, $\overline{\alpha}_{j,t}^k$ in \eqref{eq:trans_model} can be substituted as the prior in \eqref{eq:spatialupdate}, i.e.,
\begin{equation}
\label{eq:spatiotemporal}
     \alpha_{j, t}^{k} = \overline{\alpha}_{j, t-1}^k + \sum_i^N \mathcal{K}_s(x_i, x_j) y_i .
\end{equation}
Algorithm \ref{al:dynamic_semantic_mapping} consists of prediction and update steps as in a recursive Bayes filtering. For the prediction step in line \algref{al:dynamic_semantic_mapping}{alg:pred}, we apply the temporal transition model with the query point's flow estimate ${v}_{j, t-1}$. The prediction step with BACC enables removal of traces left by moving objects ``exiting'' voxels. For example, if a car was in motion at time stamp $t-1$ and moving out from a voxel $j$, calculating $v_{j, t-1}^{\text{car}}$ ensures that the map maintains confidence about a static class such as ``road'' $\overline{\alpha}_{j, t}^{\text{road}}$ and decreases confidence about the car class $\overline{\alpha}_{j, t}^{\text{car}}$. Additionally, the FORC in our algorithm facilitates the ``entry'' of dynamic objects into previously encountered areas in the map by reducing overconfidence in ``static'' and ``free'' classes. With the update step in line \algref{al:dynamic_semantic_mapping}{alg:correct}, incoming spatial and semantic training data $(\mathcal{X}_t, \mathcal{Y}_t)$ is incorporated.
\begin{algorithm}[t]
\caption{Dynamic Semantic Mapping}
\label{al:dynamic_semantic_mapping}
\small 
\begin{algorithmic}[1]
% \State \lu{Please correct this if something is wrong}
% \prompt{Shwarya: do you think the usage of $k_v$ / $k_s$ will be confusing to readers who are unfamiliar with the work? I'm fine with it personally.}\mgj{More clarity is always good. It's better to define everything of course and leave comments to reiterate what they are. It's not easy to remember everything. A good guide is to think if the flow of the paper is smooth. If readers need to go back and forth to check things and remember that's annoying. Add descriptive comments and equations numbers where relevant (as comments).}
\State \textbf{Input:} Training data: $(\mathcal{X}_{t}, \mathcal{Y}_{t}, \mathcal{V}_{t})$; \textbf{Query point}: $x_j$; \textbf{Previous state estimate} ${\alpha}_{j,t-1}$; \textbf{Flow estimate} ${v}_{j, t}$
% \Statex
\Procedure{Combined Posterior Update}{}\label{alg:proc2}
\For{$k=1,...,K$} \Comment{Prediction step}

% \Statex Prediction step
    \State $\alpha_{j, t}^k \gets \exp( - (v_{j,t-1}^k)^2 )\alpha_{j, t-1}^k$      \Comment{Incorporate motion}\label{alg:pred}
    % \State $\overline{\alpha}_{j, t}^k \gets \exp( - (v_{j,t}^k)^2 )\alpha_{j, t-1}^k$      \Comment{Incorporate motion}\label{alg:pred}
    % \State $\alpha_{j, t}^k \gets \overline{\alpha}_{j, t}^k$ \Comment{Set new hyperparameters}
    
% \Statex Update Step
\For{each $(x_i, y_i) \in (\mathcal{X}_{t}, \mathcal{Y}_{t})$} \Comment{Update Step}
    \State $w_s \gets \mathcal{K}_s(x_i, x_j)$ \Comment{Weight of $x_i$ on observation}
    \State $\alpha_{j, t}^k \gets \alpha_{j, t}^k + w_s [y_i = k]$ \label{alg:correct}
\EndFor
\EndFor
\State \textbf{return} ${\alpha}_{j, t}$ 
\EndProcedure
% \Statex
\end{algorithmic}
\end{algorithm}
% \begin{algorithm}[t]
% \caption{Dynamic Semantic Mapping}
% \label{al:dynamic_semantic_mapping}
% \small 
% \begin{algorithmic}[1]
% \Procedure{Posterior Update}{$\mathcal{D_{t}}, x_j, {\alpha}_{j,t-1}, {v}_{j, t}$}\label{alg:proc2}
% \For{$k=1,...,K$} \Comment{Prediction step}
% % \Statex Prediction step
%     \State $\alpha_{j, t}^k \gets \exp( - (v_{j,t-1}^k)^2 )\alpha_{j, t-1}^k$      \Comment{AR Motion Model}\label{alg:pred}
%     \State{$\alpha_{j, t}^{k} = \alpha_{j, t}^k + \sum_i^N \mathcal{K}_s(x_i, x_j) y_i$} \Comment{Semantic Update}\label{alg:correct}
% \EndFor
% \State \textbf{return} ${\alpha}_{j, t}$ 
% \EndProcedure
% % \Statex
% \end{algorithmic}
% \end{algorithm}

\section{Quantitative Evaluation for Dynamic Mapping}\label{sec:eval}

%Motivate why we need this metric
Typically, dynamic and semantic mapping methods that operate on stereo images re-project the map onto the image plane, and evaluation is done based on pixel-wise semantic segmentation of the image~\cite{kochanov2016scene,barsan2018robust}. Suppose one uses the same quantitative metric to evaluate the entire scene's geometric-semantic reconstruction accuracy, \textcolor{black}{this would fail} to capture the ``complete'' scene --- i.e., how well the map can represent portions of the environment that are not \textcolor{black}{reflected on the evaluation image, e.g., free cells in the map}. This problem is significant for evaluating \textcolor{black}{dynamic maps as free space in the environment might be mis-classfied as occupied due to artifacts.} Therefore, we propose a \emph{querying framework} for dynamic semantic occupancy mapping that considers both the ``complete'' scene and scene dynamics for the map evaluation.

\begin{figure}[t]
    \centering
    \includegraphics[width=0.5\textwidth,trim={0cm 0 0 3mm},clip]{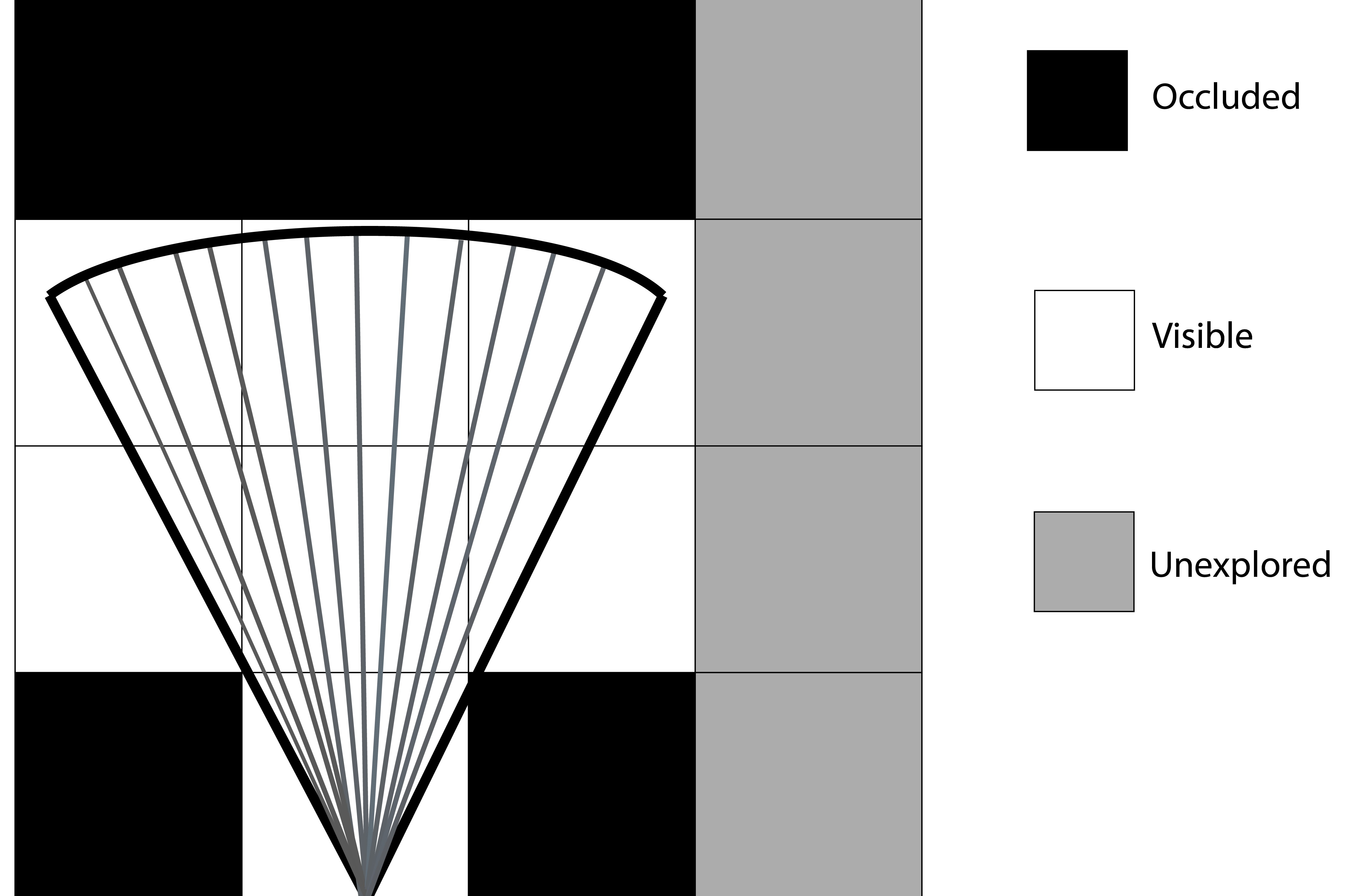}
    \caption{An illustration of the ground truth model as viewed from the robot's perspective. Rays pass through the \textcolor{black}{areas} that are observed by the robot. \textcolor{black}{Areas} that are unseen by the robot in the current scan could have been previously observed but are currently occluded. Alternatively, there could also be \textcolor{black}{areas} unexplored by the robot but present in the ground-truth model created by the multi-view dataset.}
    \label{fig:gtmap}
\end{figure}

%\textcolor{red}{[Lu: $\mathcal{M}$ is used for the set of moving classes in previous context. Maybe change one of the notations to avoid confusion.]}
\textcolor{black}{Let $\mathcal{M}$ be the map we are building to represent an environment. Let the corresponding ground truth model of the environment be denoted by $\mathcal{G}.$ The ground truth model could be  sensor data post-processed with correct labels or as a map representation --- e.g. a semantically-labelled point cloud, a set of RGB-D images, a heightmap, etc.} Let us assume that the rays from our current scan \textcolor{black}{intersect voxels that are ``observed'' by the robot (marked as ``Visible'' in Figure~\ref{fig:gtmap}). We model this "Visible" portion of $\mathcal{G}$ with $\mathcal{M}_v$ i.e., voxels in the map currently being ``observed'' by the robot.} The portion of the environment that is not seen in the current scan \textcolor{black}{could then either be previously explored or still unexplored. If some portion of the environment is ``Unexplored'' as in Figure~\ref{fig:gtmap}, there would be no voxel created in $\mathcal{M}$ for that portion. Otherwise, if the voxel was previously explored and is not visible to the robot in the current scan, we consider it ``Occluded'' as shown in Figure~\ref{fig:gtmap}. We denote the voxels in $\mathcal{M}$ that represent the occluded portion of the environment as $\mathcal{M}_o$. Our map $\mathcal{M}$ is thus comprised of voxels that are visible in the current scan ($\mathcal{M}_v$) and voxels that are not ($\mathcal{M}_o$).}

To perform a quantitative comparison on the semantic scene representation between our model $\mathcal{M}$ and any ground truth $\mathcal{G}$, we \textcolor{black}{build upon} two different map query \textcolor{black}{frameworks} introduced in~\cite{seitz}: accuracy and completeness. \textcolor{black}{In both metrics, we assess the intersection between $\mathcal{M}$ and $\mathcal{G}$. However, in map accuracy, we evaluate each element in $\mathcal{M}_v$ against $\mathcal{G}$, and in map completeness each element of $\mathcal{G}$ against $\mathcal{M}_v \bigcup \mathcal{M}_o$.}

\subsection{Map Accuracy} 
\textcolor{black}{The accuracy of the map, as the name suggests, quantifies how \emph{correct} the visible metric-semantic representation $\mathcal{M}$ is when compared with the true value $\mathcal{G}$ of the environment. As this work pertains to semantic occupancy mapping, ``correctness'' is specifically the semantic classification accuracy.}

\textcolor{black}{For each element (in our case, voxel) $m \in \mathcal{M}_v$, we generate the corresponding ground truth $g_m \in \mathcal{G}$ that is the ``closest semantic neighbor'' to $m$. In practice, $g_m$ will be the nearest element in $\mathcal{G}$ to $m$ in metric space and also, most representative of the semantic category $m$ could belong to. For example, if $\mathcal{G}$ is a point cloud there may be many points residing within $m$. The most representative semantic category is then the majority semantic label of points within voxel $m$.} 

\textcolor{black}{The semantic prediction for voxel $m$ can then be evaluated against that of $g_m$. Details about how $g_m$ can be generated from sensor data for single- and multi-view data sets are discussed in Section \ref{sec:quant}. If we are comparing the map accuracies of mapping methods with different map representations (e.g. uniform resolution voxel map versus point cloud map with a non-uniform point distribution), however, the query elements \mbox{$m \in \mathcal{M}_v$} could cover metric space differently. Therefore, map accuracy is more suited for comparing maps under the same representation (e.g. both are uniform resolution voxel maps). Consequently, we compare Semantic-BKI (S-BKI) and Dynamic-BKI (D-BKI) using the same map representation, and compute the precision, recall, and Jaccard scores across all classes.}

%In practice, both $\mathcal{G}$ and $\mathcal{M}_v$ must have the same resolution and origin so $g_m$ and $m$ have the same voxel center $\in \mathbb{R}^3$.

\begin{figure*}[t]
\centering
% \captionsetup{justification=centering}
\begin{minipage}{\textwidth}
\begin{minipage}{.6\textwidth}
\includegraphics[width=\textwidth]{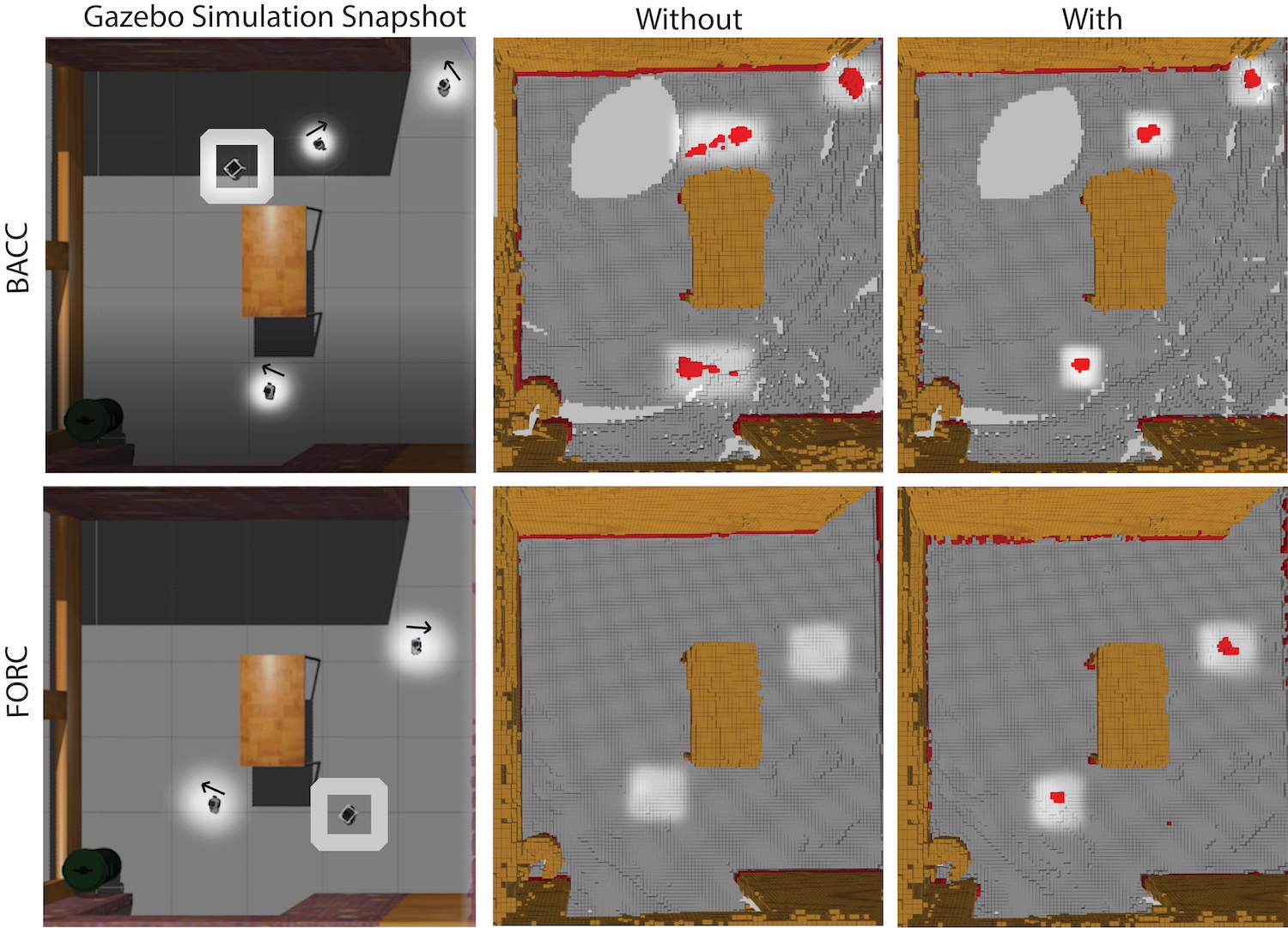}
%    \cline{1} \rotatebox{90}{FORC} & \multicolumn{3}{|c|}{}\\
    % \includegraphics[width=0.23\textwidth,trim={1.2cm 0 0 2cm},clip]{media/forward_base.jpg}&
    % \includegraphics[width=0.23\textwidth,trim={0 0.5cm 0 1cm},clip]{media/FWO.jpg}&
    % \includegraphics[width=0.23\textwidth,trim={0.5cm 1cm 0 1cm},clip]{media/FW.jpg}%\\
    % \hline
\end{minipage}
%\hspace{cm}
\begin{minipage}{.35\textwidth}
\caption{Ablation Studies with Gazebo Simulation. The images in the top row demonstrate the functionality of BACC, while the bottom row that of FORC. \textbf{Gazebo Simulation Snapshot:} In this column, we show the top view of the gazebo simulation. The ego robot is demarcated within a white square, while other moving Turtlebots are highlighted and marked with their orientation. \textbf{Without:} The images in the middle column show the global map made without specific modules. Traces are left in the map where Turtlebots were present in previous time steps without BACC. Without FORC, the map fails to represent the other two Turtlebots in the map. \textbf{With:} The right-most column shows the the global maps constructed with our approach. Minimal traces are left in the map, and the Turtlebots are in their correct locations.}
\label{fig:ablation}
\end{minipage}
\end{minipage}
\end{figure*}

\subsection{Map Completeness}
Completeness of the map pertains to how much of the environment, including both visible and occluded regions $\mathcal{M}_v \bigcup \mathcal{M}_o$, is reconstructed correctly. \textcolor{black}{For instance, if we want to evaluate a portion of the environment that was previously observed, but is not currently visible, ``completeness'' can be used to ascertain whether the map is able to represent the environment correctly. This is because we sample the environment to query the map, rather than the other way around. If we use multiple views of the environment to obtain ground truth $\mathcal{G}$ for portions of the environment that are currently occluded in the map $\mathcal{M}_o,$ it is possible to include $\mathcal{M}_o$ into completeness. Another advantage of using map completeness as a metric is that
we can compare the map inference performance across methods with different map representations.}

\textcolor{black}{We pick an element $g \in \mathcal{G}$ and find its ``closest semantic neighbor'' $m_g \in \mathcal{M}$. Again, $m_g$ would be the nearest element to $g$ in metric space and most representative semantic category that $g$ could belong to. In practice, we seek the voxel in which the element $g$ falls.} If the metric distance between $g$ and $m_g$ is greater than a certain margin, we consider that $g$ is currently unexplored by the robot \textcolor{black}{, and exclude} these pairs in the evaluation. These kinds of space are shown in the rightmost column in Figure~\ref{fig:gtmap}.

\textcolor{black}{In addition, to keep the evaluation relevant to dynamic semantic mapping, we treat static and dynamic objects differently when calculating completeness.}
\begin{enumerate}

\item \textbf{If $g$ is static.} \textcolor{black}{As $g$ is static, $g$ could not move irrespective of whether the robot has observed it. As a result, we evaluate semantic accuracy for any nearest neighbor $m_g \in \mathcal{M}$.}
\item \textbf{If $g$ is dynamic.} As only dynamic objects in $\mathcal{M}_v$ are currently seen by the robot, we evaluate semantic accuracy for all $m_g \in \mathcal{M}_v$. We do not evaluate on $\mathcal{M}_o$ as these voxels in the map are occluded and could have a different state from the last time observed by the robot.
\end{enumerate}

\subsection{Auxiliary Task: Semantic Segmentation}

\textcolor{black}{Semantic segmentation of a point cloud is another task that can be performed with an existing map model $\mathcal{M}$, and we can consider the map's performance on this auxiliary task for additional evaluation. With semantic mapping, we can inherently fuse multi-frame measurements and improve semantic classification accuracy as in S-BKI\cite{gan2019bayesian}. The querying method for a single scan is simple - we pick each point in $\mathcal{D}_t$ already inserted into the map $\mathcal{M}$ at time stamp $t$ and check which voxel it falls inside. $\mathcal{D}_t$ contains $\mathcal{Y}_t$, which are the semantic label predictions corresponding to $\mathcal{X}_t$. As discussed in Section~\ref{sec:preliminaries}, this is typically obtained from a neural network. The semantic category of the voxel becomes the prediction from our model $\mathcal{M}$. Both of these can then be compared with the ground-truth, semantically-annotated point cloud - that is typically provided in data sets such as SemanticKITTI~\cite{behley2019semantickitti}. This comparison can show whether semantic segmentation predictions can be improved through smoothing \cite{gan2019bayesian}.}
\section{Results and Discussion}
\label{sec:label}

\begin{table}[t]
\caption{Parameters for Ablation Studies.}
    \centering
    % \resizebox{\columnwidth}{!}{
\begin{tabular}{l|c}
    \toprule 
    Map resolution & 0.05 \\
    Downsampling resolution & 0.1 \\
    Free space sampling resolution & 0.5 \\
    $l_s$ & 0.15 \\
    $\sigma_s$ & 0.2 \\
    $l_1$ & 0.2 \\
    $\sigma_1$ & 50 \\ \bottomrule
    \end{tabular}
    % }
    
    % \caption{Parameters for Ablation Studies. We tune parameters for $\mathcal{K}_v$ only and adjust $\sigma_1$ according to the number of points in the point cloud. $l_s$ and $\sigma_s$ are the spatial kernel length scale and scale parameters respectively from ~\cite{gan2019bayesian}.}
    \label{tab:parameter_table}
\end{table}

\begin{table}[t]
    \centering
    \caption{Parameters for SemanticKITTI results. \textcolor{black}{Map resolution and downsampling resolution varies with Map Accuracy and Map Completeness, but the same as each other.}}
    % \resizebox{\columnwidth}{!}{
    \begin{tabular}{l|c}
    \toprule 
    Map resolution & 0.1-0.3 \\
    Downsampling resolution & 0.1-0.3 \\
    Free space sampling resolution & 100 \\
    $l_s$ & 0.1 \\
    %$\sigma_s$ & 0.2 \\
    $l_1$ & 2.5 \\
    $\sigma_1$ & 100 \\ \bottomrule
    \end{tabular}
    % }
    
    \label{tab:quantitative_parameters}
\end{table}
In this section, we first describe our experimental setup and the data sets used for evaluation. Then, we demonstrate the performance of the proposed mapping system D-BKI with qualitative results on synthetic and real data sets. Finally, quantitative results on semantic scene understanding sub-tasks using single- and multi-view data sets are presented.

% \sout{In this section, we present results evaluating the method on both simulated and real scenarios with dynamic objects. We first describe our experimental setup, and then present qualitative results on a synthetic data set generated in Gazebo and SemanticKITTI~\cite{behley2019semantickitti} data set. Quantitative results on SemanticKITTI Semantic Segmentation competition~\footnote{\href{https://competitions.codalab.org/competitions/20331}{https://competitions.codalab.org/competitions/20331}} which placed second out of 69 participants at the time of this submission are also presented. Our code will be made publicly available after receiving the final decision. }
\subsection{Experimental Setup}\label{sec:experiments}
We first describe our (i) system design choices, then elaborate on (ii) the data sets used, and lastly discuss (iii) flow estimation for each point cloud data set.
\subsubsection{System Design Choices}
In the \textcolor{black}{proposed} mapping framework, every query voxel has 6 neighbours (one on each \textcolor{black}{facet}). \textcolor{black}{For computational efficiency, only the training points within the 6 neighbouring voxels are used} in the calculation of both \eqref{eq:velocity} and \eqref{eq:velfree}. %\textcolor{red}{Only using neighboring voxels breaks the sparse kernel, as requirement is 0 if $d \geq l$.}

We choose a sparse kernel~\cite{melkumyan2009sparse} as $\mathcal{K}_v = \mathcal{K}_v^{\text{free}} = $
\begin{align}
\small 
    % \nonumber &\mathcal{K}_v^{\text{free}}(x,x') = \\
    \nonumber &\begin{cases}
        \sigma_1[(\frac{1}{3}(2 + \cos({\frac{2\pi d}{l_1}})(1 - \frac{d}{l_1}) + \frac{1}{2\pi}\sin({\frac{2\pi d}{l_1}})] & \text{if } d < l_1\\
        0 & \text{otherwise}
    \end{cases}
\end{align}
where \mbox{$d = \lVert x-x'\rVert$, $l_1 > 0$} is the length scale and $\sigma_1$ is the kernel scale parameter. %To compute $\mathcal{K}_v$ in \eqref{eq:velocity}, we use an isotropic Mat\'ern kernel~\cite{stein1999interpolation}, \mbox{$\mathcal{K}_v(x,x') = \sigma_2[(1 + \frac{\pi d}{2l_2}) e^{-\frac{\pi d}{2l_2}}]$}, where $d = \lVert x-x'\rVert$, $l_2 > 0$ the length scale and $\sigma_2$ the kernel scale parameter.

Typically, the kernel length scale in $\mathcal{K}_v$ and $\mathcal{K}_v^{\text{free}}$ is chosen with respect to the resolution of the map being built as it controls how much influence a point in a neighbouring \textcolor{black}{voxel} has. In our experiments, $l_1$ is \textcolor{black}{greater than} the map resolution and $\sigma_1$ can be set once in the beginning according to the size of the point set and free-space sampling rate. Lastly, in Algorithm \ref{al:sceneflow}, we \textcolor{black}{implement $f(.,.)$ as} a moving average filter in line \algref{al:sceneflow}{alg:filter}.

\subsubsection{Data Sets and Benchmarks}\label{sec:gaz_env}
We use point cloud-based data sets that contain only positional information. However, our method is amenable to any point cloud data with intensity, colour or other fields. Additionally, it can also be applied to depth camera data, of which larger data sets for training and evaluation exist.
\begin{figure}[ht]
    \centering
    \subfloat[Sequence 01]{
        \includegraphics[width=\linewidth,trim={0 0 0 0},clip]{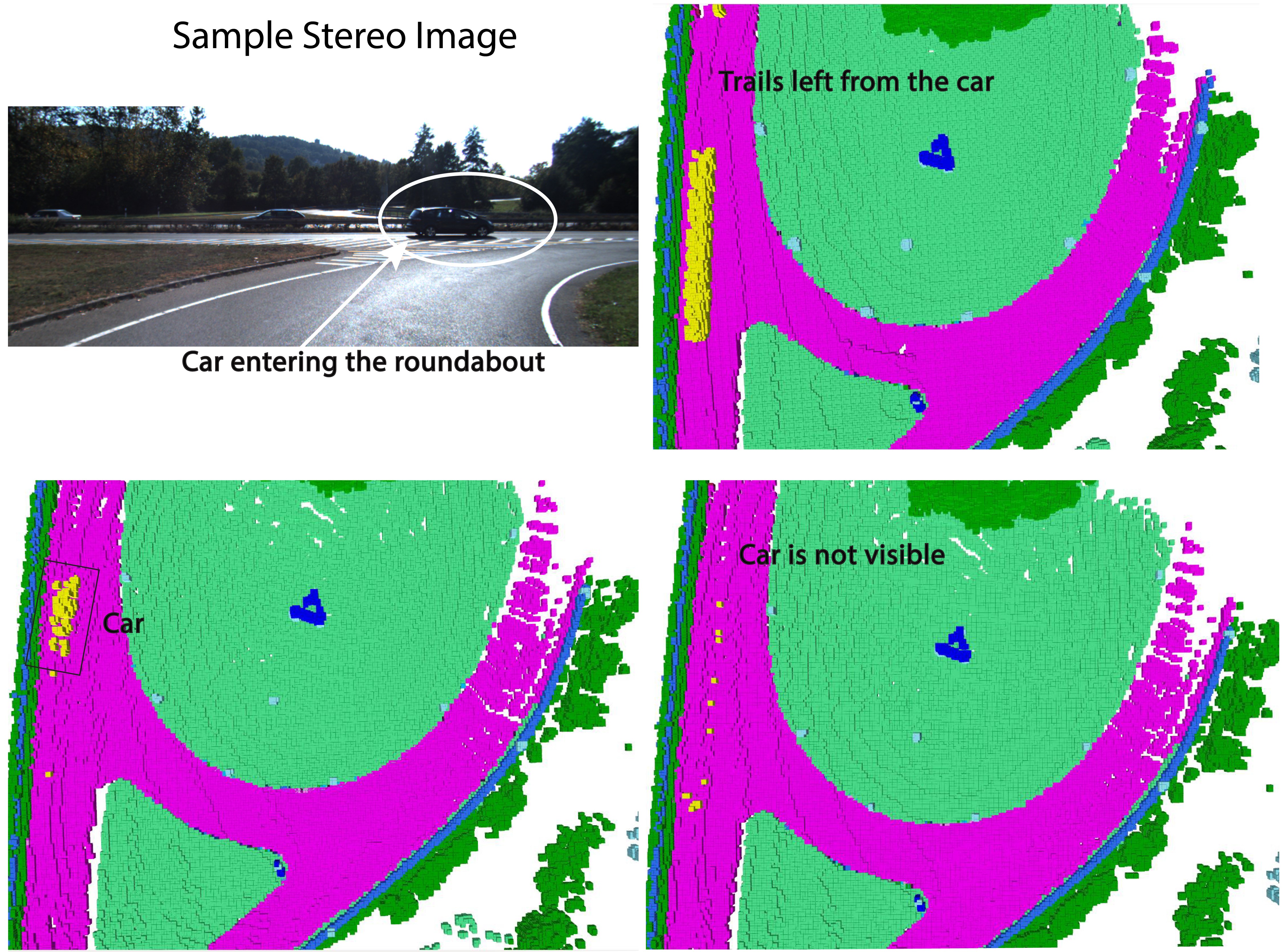}
        \label{fig:se1_1_qualitative}
    }
    \\
    \vspace{-2mm}
    \begin{tabular}{m{\linewidth}}
    \\
    \hline
    \end{tabular}
    \\
    \subfloat[Sequence 04]{
        \includegraphics[width=\linewidth,trim={0 0 0 0},clip]{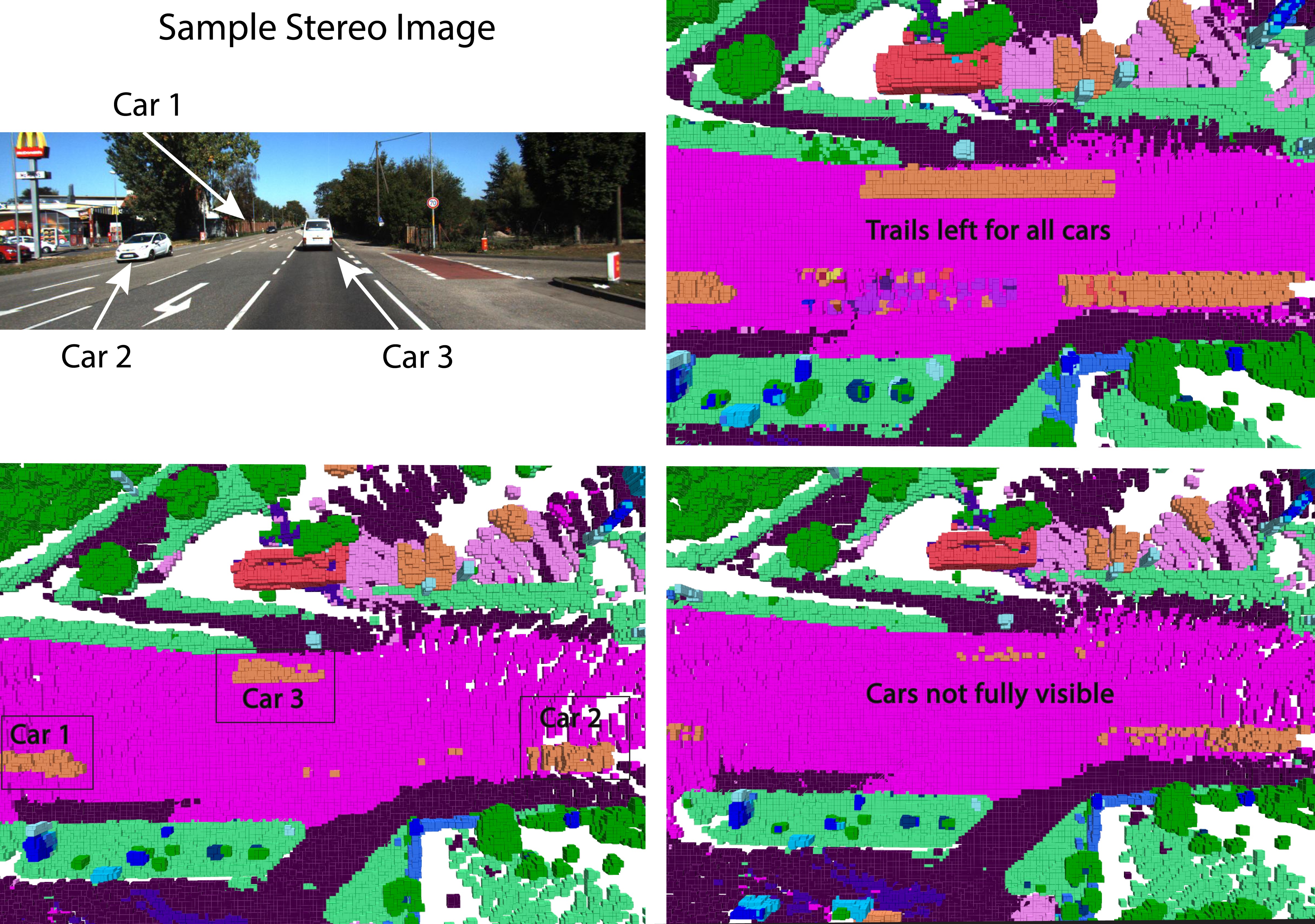}
        \label{fig:seq_4_qualitative}
    }
    \caption{Qualitative results on Sequences 01 (top) and 04 (bottom) of SemanticKITTI. Four images are shown for both frames, which include (Top Left:) Right stereo image corresponding to one of the scans. Note that the image is included for validation, however we strictly perform map update from LiDAR when generating results. (Top Right:) S-BKI mapping without any free space sampling. Trails are left where cars passed over. (Bottom Right:) S-BKI with free space sampling. After a few scans, the map becomes overconfident about the presence of free cells and fails to incorporate dynamic objects in the map. (Bottom Left:) D-BKI with free space sampling. The cars are tracked with minimal traces.} 
    \label{fig:qualitative_kitti}
\end{figure}

\begin{figure*}[ht]
    \centering
    \subfloat{
        \includegraphics[width=0.5\textwidth,trim={0 0 0 0},clip]{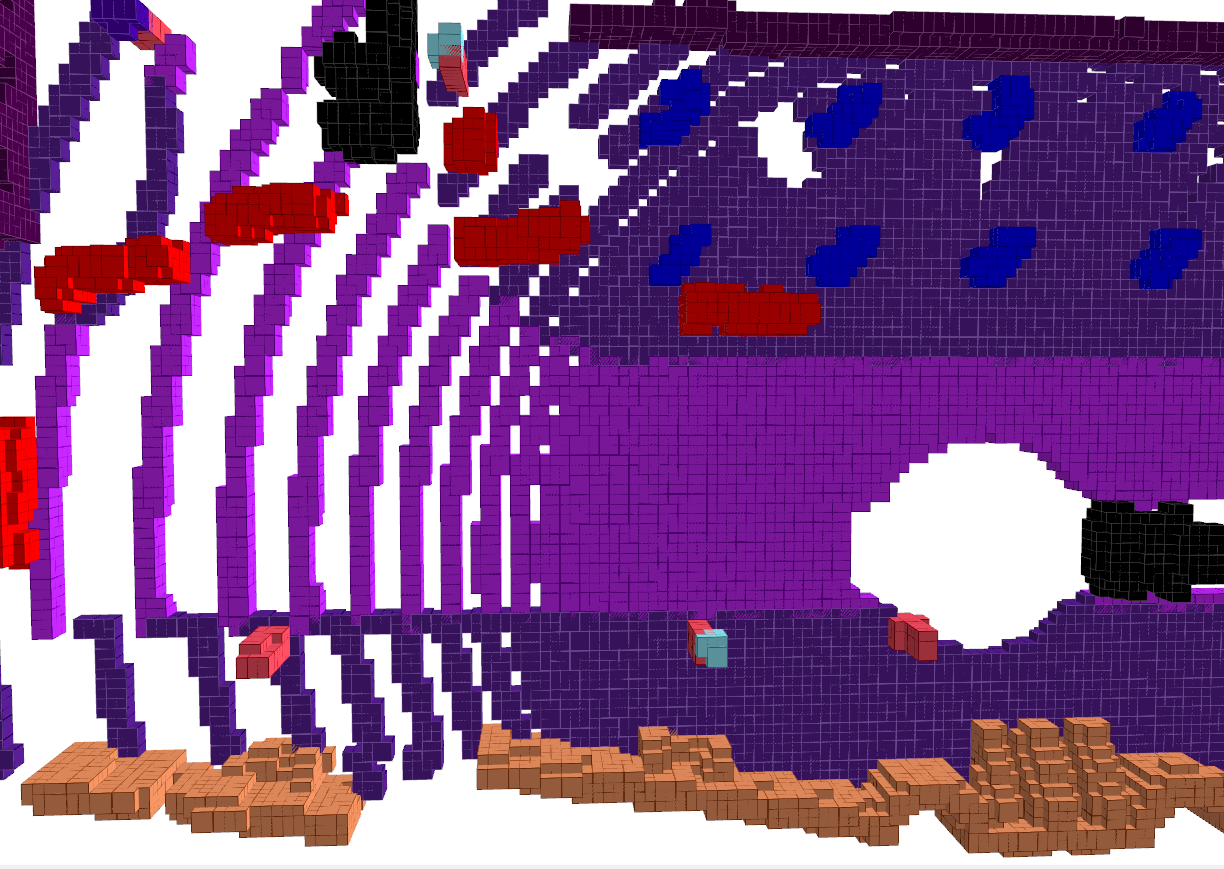}
        \label{fig:carla_static}
    }\vline
    \subfloat{ 
    \includegraphics[width=0.5\textwidth,trim={10 10 10 20},clip]{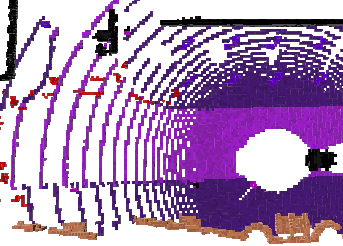}
        \label{fig:carla_bl}
    }\\
    \subfloat{ 
    \includegraphics[width=0.5\textwidth,trim={0 0 0 0},clip]{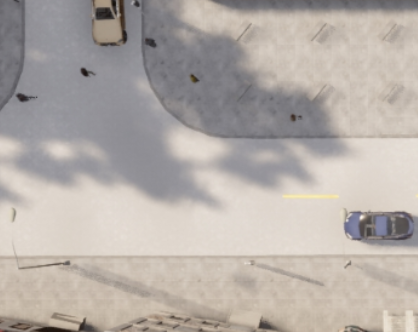}
        \label{fig:carla_bev}
    }\vline
    \subfloat{
        \includegraphics[width=0.5\textwidth,trim={50 0 80 50},clip]{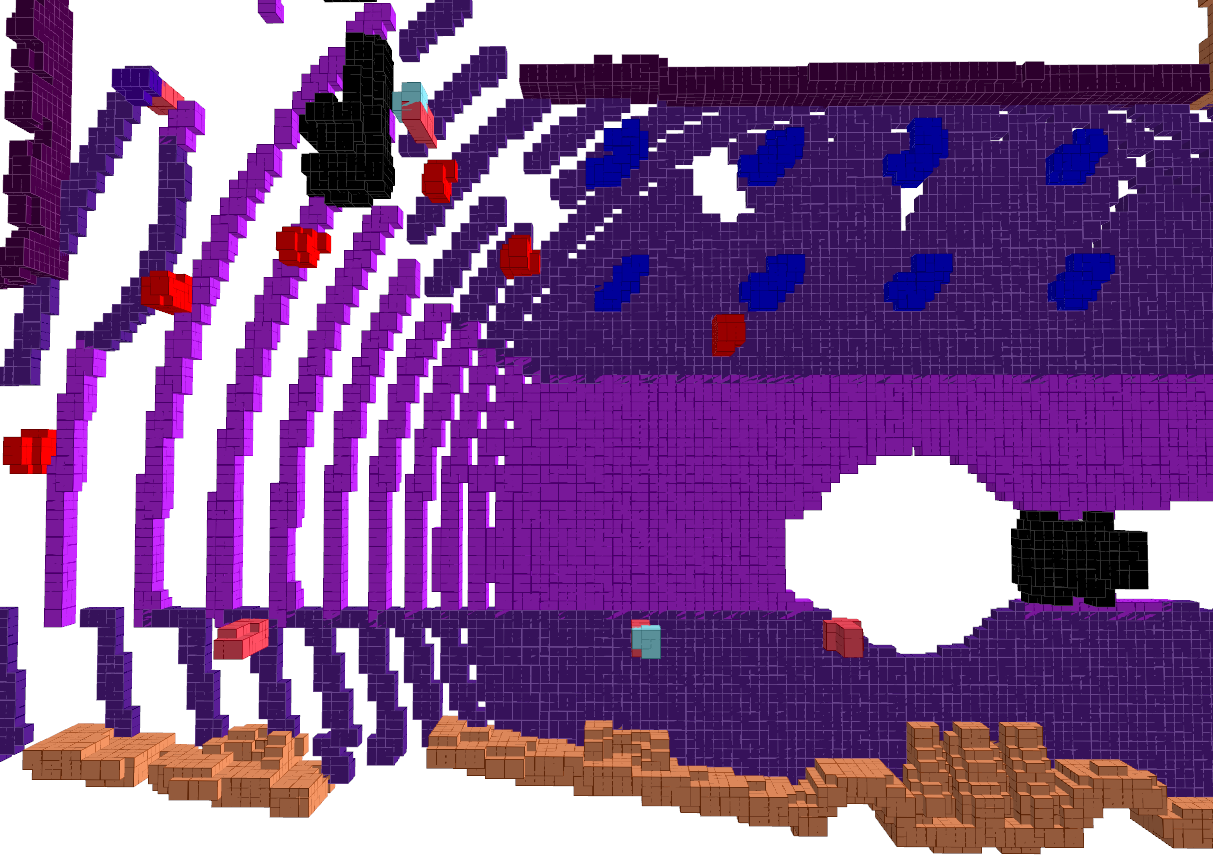}
        \label{fig:carla_dynamic}
    }
    
    \caption{Qualitative results on S-BKI (top left), Kochanov et al. (top-right) and D-BKI (bottom-right) of the CARLA data set. This scene (bottom-left) showcases a static ego-car (in dark-black) parked on the road while pedestrians walk around within its sensing range. In the map images, the red blobs are pedestrians walking on the street and the black rectangles are other cars (whether parked or moving). One can see that D-BKI "fills" in the shape of the car more than Kochanov et al. (as in the car on the top left) and does not leave trails of the walking pedestrians as with S-BKI. For Kochanov et al., there seems to be a more sparse representation for pedestrians due to their unique shape than D-BKI.} 
    \label{fig:qualitative_carla}
\end{figure*}

\subsubsection*{Gazebo Simulation Environment} To create an indoor synthetic data set, a Gazebo simulation environment was set up with multiple Turtlebots exploring a house. We mounted one robot (the ego-robot) with an omni-directional block laser scanner for data collection in the form of point clouds with positional information only. To simulate dynamic objects in the environment, we have other three Turtlebots exploring the same house. Using a reactive planner, the robots avoid each other and obstacles in the environment. The collected data is processed using Point Cloud Library (PCL) \cite{pcl} and annotated based on height into three \textcolor{black}{semantic classes} - floor, robot and miscellaneous objects including walls and cabinets. The scene flow for each scan is computed with FlowNet3D \cite{flownet3d}. 

\subsubsection*{SemanticKITTI Data Set} The SemanticKITTI data set~\cite{behley2019semantickitti} is a large-scale real driving data set based on the KITTI Vision Benchmark \cite{kittiBenchmark} where \textcolor{black}{semantically-annotated LiDAR scans} and camera poses are provided for all sequences. Camera poses are estimated with SuMa~\cite{behley2018efficient}, and semantic annotations \textcolor{black}{for each LiDAR scan} are generated by RangeNet++\cite{milioto2019rangenet++}. Additional labels are provided to distinguish static objects from dynamic objects, such as person and moving-person. There are 22 sequences, out of which 11 sequences are provided with ground truth labels for training (00-07), validation (08) and testing (09-10). Sequence 11-21 do not come with ground truth semantic labels, but can be evaluated in a public leaderboard over the mean Intersection-over-Union (mIoU) metric. Since the most reliable ground truth model in this single-view data set is the semantically-labeled point cloud, we use them to generate a ground truth model $\mathcal{G}$.

\subsubsection*{CARLA Data Set}
\textcolor{black}{To evaluate the map completeness of the proposed dynamic semantic mapping, a reliable ground truth model of the environment including free space is needed. As real data sets collected using a single-view sensor (such as SemanticKITTI) usually do not have sufficient measurement coverage to fully recover the underlying environment model, we leverage a simulation environment CARLA~\cite{Carla}.}
% To generate a reliable ground truth model of the scene for evaluation of the global map, we obtain multiple views of the same scene and fuse the data into a \emph{voxel map}. The voxel centers in the voxel map act as query points in the \textcolor{black}{map} completeness experiments.

We generate a synthetic \textcolor{black}{multi-view} scene completion data set sequence from the CARLA~\cite{Carla} simulator. The methodology for its creation is available publicly in~\cite{MotionSC}. We generate \textcolor{black}{ground-truth environment models by uniformly distributing multiple LiDAR sensors around the ego vehicle, effectively obtaining a 3D Monte Carlo sampling of the world which is i.i.d. with respect to time. The simulation environment also provides ground-truth scene flow (velocity) and semantic labels for each point.} Free space observations are obtained by linearly interpolating along all points at a fixed interval of 1.5 meters. \textcolor{black}{Ground truth point clouds with semantic labels are then fused into a semantically annotated ground truth voxel model $\mathcal{G}$ with 0.3 meter resolution. The voxel centers in the ground truth model act as query points in the completeness experiments.} This approach is similar to the SemanticKITTI~\cite{behley2019semantickitti} scene completion data set; however, it has no traces from dynamic objects and fewer occlusions due to sampling from multiple sensors at the \emph{same} time. %The voxelized world dimensions are from (-30 m, -30 m, -2.5 m) to (30 m, 30 m, 2.5 m) at a 0.1 meter resolution per axis, on a left handed Euclidean coordinate system. %Free space observations are sampled with 1.5 meters spacing between each observation. In the event a voxel has multiple observations, the voxel is assigned the semantic class with the most observations.  

%\todo{Add citation for GitHub page}

%Voxels are dived into two groups for quantitative evaluation: visible and occluded. Visible voxels are determined to be those observed by the onboard ego LiDAR sensor, including free space. Occluded voxels are all voxels unobserved by the onboard ego sensor. By dividing the cells, we gain insight into the completeness (occluded) and accuracy (visible) performance of each algorithm.

% We use the same ray-tracing approach to sample free space from SemanticKITTI~\cite{behley2019semantickitti} LiDAR scans. For each scan, free space is obtained by ray-tracing backward from all points in the point cloud. Since there is only a single sensor, we do not voxelize points and instead query the exact point location. When there are twenty sensors, this approach is less feasible as it would require querying millions of points per frame. Free-space sampling distance is a constant 1.5 meters in both CARLA and KITTI data sets. Links to both data sets may be found in our GitHub. 

\subsubsection{Flow Estimation}
% As discussed in Algorithm \ref{al:sceneflow}, we compute motion fields for each query point $X_j$ in the map. 
To obtain the corresponding flows \textcolor{black}{$\mathcal{U}_t$} of a point cloud $\mathcal{X}_t$, we \textcolor{black}{choose} a state-of-the-art deep learning architecture based on PointNet++~\cite{PointNetPP} -  FlowNet3D~\cite{flownet3d}: \textcolor{black}{a} supervised method based on PointNet++ which estimates scene flow between two successive point clouds $\mathcal{X}_{t-1}$ and $\mathcal{X}_t$.
%;\sout{2) TLFPAD \cite{tlfpad} : self-supervised method based on PointNet++ \cite{PointNetPP} which incorporates temporal information from four point clouds $\mathcal{X}_{t-3:t}$. Predicted scene flow $\mathcal{V}_t$ is compared with the next point cloud $\mathcal{X}_{t+1} = \mathcal{V}_t + \mathcal{X}_t$.}

Typically, implementations for scene flow estimation train on the XYZRGB fields, i.e., the point cloud includes both the position and colour information. We trained an adapted version of FlowNet3D on the KITTI 2015 Scene Flow data set~\cite{menze2015object} and the FlyingThings driving data set~\cite{mayer2016large} by only including position information for training. After obtaining \textcolor{black}{$\mathcal{U}_t$} from the networks, we perform egomotion-compensation by subtracting the mean flow of static classes \textcolor{black}{from $\mathcal{U}_t$}.

As we only need flow for moving objects, our mapping method can be applied when any flow information for $\{x_t^q \in \mathcal{X}_t \mid \forall q \in \mathcal{Q}\}$ is available. To demonstrate the performance of the mapping framework \textcolor{black}{independent of flow estimation error}, we obtain the ground truth velocity of dynamic objects from the CARLA simulator.

\subsection{Qualitative Results}
The goal of this section is to (i) demonstrate improvements of the temporal transition model qualitatively through ablation studies, and (ii) compare the real-time map construction by semantic and dynamic BKI with the data sets described in Section \ref{sec:gaz_env}. 

% Ablation studies are conducted in the Gazebo simulation environment, and images of the global map constructed on SemanticKITTI are provided for comparison between semantic and dynamic mapping.  
% \begin{figure}
%   \includegraphics[width=\linewidth, trim={2cm 1cm 3.5cm 1.5cm},clip]{media/scene.png}
%   \caption{Simulation environment with multiple Turtlebots encountering each other in their exploration of a house\todo{remove, replace with KITTI qualitative}}
%   \label{fig:base}
% \end{figure}

% \begin{minipage}{\columnwidth}
% \begin{minipage}[t]{0.48\columnwidth}
% \makeatletter\def\@captype{table}
% \centering
% \resizebox{\columnwidth}{!}{
% \begin{tabular}{l|c}
%     \toprule 
%     Map resolution & 0.05 \\
%     Downsampling resolution & 0.1 \\
%     Free space sampling resolution & 0.5 \\
%     $l_s$ & 0.15 \\
%     $\sigma_s$ & 0.2 \\
%     $l_1$ & 0.2 \\
%     $\sigma_1$ & 50 \\ \bottomrule
% \end{tabular}}
% \caption{Parameters for Ablation Studies.}
% \label{tab:parameter_table}
% \end{minipage}
% \hfill\vline\hfill
% \begin{minipage}[t]{0.48\columnwidth}
% \makeatletter\def\@captype{table}
% \centering
% \resizebox{\columnwidth}{!}{
% \begin{tabular}{l|c}
%     \toprule 
%     Map resolution & 0.1 \\
%     Downsampling resolution & 0.1 \\
%     Free space sampling resolution & 100 \\
%     $l_s$ & 0.1 \\
%     %$\sigma_s$ & 0.2 \\
%     $l_1$ & 2.5 \\
%     $\sigma_1$ & 100 \\ \bottomrule
% \end{tabular}}
% \caption{Parameters for SemanticKITTI results.}
%     \label{tab:quantitative_parameters}
% \end{minipage}
% \end{minipage}

\subsubsection{Ablation Studies}
We perform two ablation studies to demonstrate the function and efficacy of each component of the method. These studies are aimed to qualitatively show how the global map inference performs without backward (BACC) or forward (FORC) correction. Note that our global map is colored with a gray floor, mustard walls and red Turtlebots. We annotate our ego-robot building the map with a white box around it. Holes on the floor are typically spaces where the sensor has not scanned yet. We only tune parameters (in Table~\ref{tab:parameter_table}) for $\mathcal{K}_v$. $l_s$ and $\sigma_s$ are the spatial kernel length scale and scale parameters respectively from~\cite{gan2019bayesian}.

\textbf{Without BACC:} To conduct this study, we remove scene flow aggregation for all dynamic classes by setting $v_{j,t-1}^q = 0 \mid \forall q \in \mathcal{Q}$ and observe the map as it is being built. Results are shown in the top row of Fig.~\ref{fig:ablation}. In the \emph{simulation snapshot}, we highlight the 3 Turtlebots in the environment that are in motion. \emph{Without BACC}, trails are visible behind each Turtlebot due to their motion not being considered during map-building. \emph{With BACC}, \textcolor{black}{no trails are left behind and each robot has consistently} the same size due to the incorporation of the temporal transition model.
    
\textbf{Without FORC:} Results for this experiment are shown in the bottom row of Fig.~\ref{fig:ablation}. The \emph{simulation snapshot} shows 2 moving Turtlebots in the environment. \emph{Without FORC}, the motion of these Turtlebots around \textcolor{black}{free voxels are} not considered to compute $v_{j,t-1}^{\text{free}}$ in \eqref{eq:velfree}. As a result, the prediction step for $\overline{\alpha}_{j, t}^{\text{free}}$ in \algref{al:dynamic_semantic_mapping}{alg:pred} becomes obsolete. If $\alpha_{j, t}^{\text{free}} > \alpha_{j, t}^{\text{robot}}$, then voxel $j$ will be (incorrectly) classified as a free cell. Note that the other two robots do not get incorporated into the map as $\alpha_{j, t}^{\text{free}} > \alpha_{j, t}^{\text{robot}}$ for the voxels. \emph{With FORC}, the map successfully represents the two Turtlebots.  

\subsubsection{SemanticKITTI Data Set}\label{sec:kittiqual}
We include images from sequence 1 and 4 of the SemanticKITTI data set to highlight the differences between static (S-BKI) and dynamic (D-BKI) mapping, as this is not easily captured in the semantic segmentation competition. These results are shown in Fig. \ref{fig:qualitative_kitti}, where S-BKI either discards dynamic objects over time completely or leaves them in the map depending on parameter choice. In contrast, D-BKI is able to accurately represent the moving objects without leaving long trails. Some of the parameters used to run the experiments are specified in Table \ref{tab:quantitative_parameters}.
\subsubsection{CARLA Data Set}\label{sec:carlaqual}
We show qualitative results on five different scenarios described in Fig.~\ref{fig:qualitative_carla} and the appendix section comparing our approach with various baselines in Fig.~\ref{fig:qualitative_jaywalk} and~\ref{fig:qualitative_heavy_moving}.

\subsection{Quantitative Evaluation}
\subsubsection{Semantic Mapping}\label{sec:quant}
For this sub-task, we compare the estimated map with respect to the ground truth world models $\mathcal{G}$ of the SemanticKITTI and CARLA simulator data set (described in Sec. \ref{sec:eval}). \textcolor{black}{We conduct our experiments with two querying and evaluation metrics - map completeness and map accuracy. Map accuracy is measured at the intersection of the \emph{visible} estimated map $\mathcal{M}_v$ with the $\mathcal{G}$, and evaluated at each voxel in the estimated map. \textcolor{black}{Map }completeness includes both visible ($\mathcal{M}_v$) and occluded voxels ($\mathcal{M}_o$), and is evaluated at each ground truth element. Note that there are some considerations about the ground truth world model generation, which we will discuss in detail next.}

\begin{table*}[t]
% \footnotesize
\centering
\caption{Quantitative results for Dynamic-BKI using two map evaluation methods on SemanticKITTI data set ~\cite{behley2019semantickitti} for 26 semantic classes. Comparisons are made with Semantic-BKI \cite{gan2019bayesian} for \textbf{Map Accuracy} and Kochanov et al. \cite{kochanov2016scene} for \textbf{Map Completeness}. The performance metric \textbf{mean IoU (mIoU)} is used to calculate a quantitative measure with the data collected.}
\resizebox{\textwidth}{!}{
\begin{tabular}{llllcccccccccccccccccccccccc}
% \toprule
% \multicolumn{3}{c|}{Method}
{\bf Map Evaluation Method} & 
\multicolumn{1}{l}{\bf Mapping Method}&
\cellcolor{scarColor}\rotatebox{90}{\color{white}Car} &
\cellcolor{sbicycleColor}\rotatebox{90}{\color{white}Bicycle} &
\cellcolor{smotorcycleColor}\rotatebox{90}{\color{white}Motorcycle} &
\cellcolor{struckColor}\rotatebox{90}{\color{white}Truck} & 
\cellcolor{sothervehicleColor}\rotatebox{90}{\color{white}Other Vehicle} & 
\cellcolor{spersonColor}\rotatebox{90}{\color{white}Person} &
\cellcolor{sbicyclistColor}\rotatebox{90}{\color{white}Bicyclist} &
\cellcolor{smotorcyclistColor}\rotatebox{90}{\color{white}Motorcyclist} &
\cellcolor{sroadColor}\rotatebox{90}{\color{white}Road} &
\cellcolor{sparkingColor}\rotatebox{90}{\color{white}Parking} &
\cellcolor{ssidewalkColor}\rotatebox{90}{\color{white}Sidewalk} &
\cellcolor{sothergroundColor}\rotatebox{90}{\color{white}Other Ground} &
\cellcolor{sbuildingColor}\rotatebox{90}{\color{white}Building} &
\cellcolor{sfenceColor}\rotatebox{90}{\color{white}Fence} &
\cellcolor{svegetationColor}\rotatebox{90}{\color{white}Vegetation} &
\cellcolor{strunkColor}\rotatebox{90}{\color{white}Trunk} &
\cellcolor{sterrainColor}\rotatebox{90}{\color{white}Terrain} &
\cellcolor{spoleColor}\rotatebox{90}{\color{white}Pole} &
\cellcolor{strafficsignColor}\rotatebox{90}{\color{white}Traffic Sign} &
\cellcolor{scarColor}\rotatebox{90}{\color{white} Car-Moving} &
\cellcolor{sbicyclistColor}\rotatebox{90}{\color{white} Bicyclist-Moving} &
\cellcolor{spersonColor}\rotatebox{90}{\color{white} Person-Moving} &
\cellcolor{smotorcyclistColor}\rotatebox{90}{\color{white} Motorcylist-Moving} &

\cellcolor{sothervehicleColor}\rotatebox{90}{\color{white} Other Vehicle-Moving} &
\cellcolor{struckColor}\rotatebox{90}{\color{white} Truck-Moving} &

\rotatebox{90}{\bf Average}\\ \hline 

\vspace{-2mm} \\
\multirow{2}{*}{{Map} Accuracy}
& D-BKI (Ours) & 0.611 & 0.588 & 0.658 & 0.751 & 0.699 & 0.598 & 0.222 & 0.595 & 0.646 & 0.585 & 0.593 & 0.437 & 0.819 & 0.562 & 0.727 & 0.469 & 0.513 & 0.456 & 0.594 & \textbf{0.708} & \textbf{0.695} & \textbf{0.717} & \textbf{0.627} & \textbf{0.761} & \textbf{0.714} & \textbf{0.614} \\
%& S-BKI & 0.954 & 0.681 & 0.875 & 0.940 & 0.865 & 0.714 & n/a & n/a & 0.959 & 0.788 & 0.871 & 0.451 & 0.912 & 0.803 & 0.895 & 0.723 & 0.803 & 0.754 & 0.823 & 0.929 & 0.906 & 0.772 & 0.613 & 0.793 & 0.585 & 0.800 \\ note: is this necessary anymore?
& S-BKI & 0.616 & 0.590 & 0.654 & 0.754 & 0.700 & 0.602 & \textbf{0.272} & 0.595 & 0.652 & 0.585 & 0.593 & 0.437 & 0.819 & 0.562 & 0.727 & 0.469 & 0.513 & 0.456 & 0.594 & 0.224 & 0.152 & 0.475 & 0.163 & 0.345 & 0.426 & 0.519 \\
\bottomrule
\multirow{2}{*}{{Map} Completeness}
& D-BKI (Ours) & \textbf{0.865} & 0.639 & \textbf{0.845} & \textbf{0.846} & \textbf{0.760} & \textbf{0.733} & n/a & n/a & \textbf{0.933} & \textbf{0.694} & \textbf{0.617} & \textbf{0.485} & 0.718 & \textbf{0.656} & 0.750 & 0.461 & 0.643 & \textbf{0.664} & 0.738 & 0.780 & 0.775 & 0.823 & 0.557 & 0.933 & 0.672 & \textbf{0.719}\\
& Kochanov et. al. \cite{kochanov2016scene}
& 0.848 & 0.640 & 0.810 & 0.824 & 0.721 & 0.684 & n/a & n/a & 0.915 & 0.627 & 0.598 & 0.457 & 0.721 & 0.634 & 0.747 & \textbf{0.479} & 0.639 & 0.627 & 0.738 & 0.773 & 0.771 & \textbf{0.847} & 0.549 & 0.930 & 0.670 & 0.707\\
%& 0.946 & 0.593 & 0.411 & 0.495 & 0.461 & 0.273 & 0.0 & 0.0 & 0.907 & 0.663 & 0.748 & 0.26 & 0.906 & 0.656 & 0.857 & 0.727 & 0.711 & 0.637 & 0.694 & 0.756 & 0.64 & 0.656 & 0.329 & 0.221 & 0.012 & 0.542 \\
\bottomrule
\end{tabular}
}
% \squeezeup
\label{tab:single_view_comp}
\end{table*}

\begin{table}[t]
% \footnotesize
\centering
\caption{\textcolor{black}{Quantitative results for Dynamic-BKI and Semantic-BKI using the map evaluation method \textbf{Map Accuracy}} on the CARLA data set. \textcolor{black}{The data collected is evaluated with two performance metrics} - \textbf{Precision} and \textbf{Recall.}}
\resizebox{0.5\textwidth}{!}{
\begin{tabular}{llllcccccccccccccccccccccccc}
% \toprule
% \multicolumn{3}{c|}{Method}
{\bf {Metric}} & 
\multicolumn{1}{l}{\bf Mapping Method}&
\cellcolor{smotorcyclistColor}\rotatebox{90}{\color{white}Free Space} &
\cellcolor{struckColor}\rotatebox{90}{\color{white}Vehicle} &
\cellcolor{spersonColor}\rotatebox{90}{\color{white}Pedestrian} &
\cellcolor{sroadColor}\rotatebox{90}{\color{white}Road} &
\cellcolor{sbuildingColor}\rotatebox{90}{\color{white}Building} & 
\cellcolor{sothervehicleColor}\rotatebox{90}{\color{white}Sidewalk} & 

\cellcolor{scarColor}\rotatebox{90}{\color{white}Traffic Sign} &
\rotatebox{90}{\bf Average}\\ \hline 

\vspace{-2mm} \\
\multirow{2}{*}{Precision}
& D-BKI (Ours) & 99.53 & \textbf{85.99} & \textbf{85.30} & 91.76 & 95.78 & 86.13 & 87.16 & \textbf{90.23} \\
& S-BKI & 99.57 & 63.10 & 24.94 & 91.42 & 95.35 & 87.53 & 87.16 &  78.44 \\
\bottomrule
\vspace{-2mm} \\
\multirow{2}{*}{Recall}
& D-BKI (Ours) & \textbf{93.17} & 97.13 & 92.89 & 98.81 & 98.94 & 98.67 & 98.67 & \textbf{96.89} \\
& S-BKI & 86.97 & 97.53 & 95.50 & 98.23 & 98.89 & 98.43 & 98.69 & 96.33  \\
\bottomrule
% \vspace{-2mm} \\
% \multirow{2}{*}{Jaccard Score}
% & S-BKI &  & & & & &  \\
% & D-BKI (Ours) & & & & & & \\
% \bottomrule
\end{tabular}
}
% \squeezeup
\label{tab:carla_acc}
\end{table}

\begin{table}[t]
% \footnotesize
\centering
\caption{\textcolor{black}{Quantitative results for Dynamic-BKI and Kochanov et. al \cite{kochanov2016scene} using the map evaluation method \textbf{Map Completeness}} on the CARLA data set. \textcolor{black}{The data collected for the visible map $\mathcal{M}_v$ is evaluated with two performance metrics} - \textbf{Precision} and \textbf{Recall}. \textcolor{black}{The data collected for the occluded map $\mathcal{M}_o$ is evaluated with} \textbf{mean IoU (mIoU)}.}
\label{table:completeness}
\resizebox{0.5\textwidth}{!}{
\begin{tabular}{llllcccccccccc}
% \toprule
% \multicolumn{3}{c|}{Method}
{\bf Indicator} & 
\multicolumn{1}{l}{\bf Mapping Method}&
\cellcolor{smotorcyclistColor}\rotatebox{90}{\color{white}Free Space} &
\cellcolor{struckColor}\rotatebox{90}{\color{white}Vehicle} &
\cellcolor{spersonColor}\rotatebox{90}{\color{white}Pedestrian} &
\cellcolor{sroadColor}\rotatebox{90}{\color{white}Road} &
\cellcolor{sbuildingColor}\rotatebox{90}{\color{white}Building} & 
\cellcolor{sothervehicleColor}\rotatebox{90}{\color{white}Sidewalk} & 
\cellcolor{scarColor}\rotatebox{90}{\color{white}Traffic Sign} &
\rotatebox{90}{\bf Average}\\ \hline 

\vspace{-2mm} \\
\multirow{2}{*}{Visible Map: Precision}
%& S-BKI & 99.7 & 83.6 & 21.31 & 84.87 & 93.06 & 87 & 60.86 \\
& D-BKI (Ours) & \textbf{99.66} & 88.59 & \textbf{85.68} & 83.24 & 93.71 & 88.99 & 72.7 & 87.08 \\
& Kochanov et al. & 98.60 & \textbf{89.94} & \textbf{85.34} & \textbf{88.59} & \textbf{94.01} & \textbf{91.86} & \textbf{80.44} & \textbf{89.82}\\
\bottomrule
\vspace{-2mm} \\
\multirow{2}{*}{Visible Map: Recall}
%& S-BKI & 96.35 & 98.29 & & & &  \\
& D-BKI (Ours) & 96.62 & \textbf{89.87} & \textbf{85.92} & \textbf{96.71} & \textbf{97.20} & \textbf{97.94} & \textbf{96.56} & \textbf{94.26}  \\
& Kochanov et al.  & \textbf{98.46} & 74.80 & 78.96 & 91.47 & 96.92 & 92.79 & 89.32 & 88.96 \\
\bottomrule
\vspace{-2mm}\\
\multirow{2}{*}{Occluded Map: mIoU}
& D-BKI (Ours) & \textbf{96.18} & - & - & \textbf{51.18} & \textbf{60.16} & \textbf{64.31} & 29.47 & \textbf{62.30} \\
& Kochanov et al. & 94.73 & - & - & 41.29 & 57.12 & 43.51 & \textbf{36.63} & 54.65\\
\bottomrule

% \vspace{-2mm} \\
% \multirow{2}{*}{Jaccard Score}
% & S-BKI &  & & & & &  \\
% & D-BKI (Ours) & & & & & & \\
% \bottomrule
\end{tabular}
}
% \squeezeup
\label{tab:carla_comp}
\end{table}

We compare our algorithm, D-BKI, against the static semantic mapping baseline S-BKI \cite{gan2019bayesian}, and a scene-propagation-based dynamic semantic mapping algorithm by Kochanov et al. \cite{kochanov2016scene}. Although \cite{kochanov2016scene} presents results on building voxel maps with stereo images, the approach is general and performs semantic segmentation and scene flow estimation to incorporate into the mapping pipeline later. We re-implemented their approach and tuned it to generate results on LiDAR point clouds. As their approach updates semantic and occupancy probability separately, we performed free space sampling to provide this extra information.

\noindent \par \textbf{Single-view data set}: For a single-view data set, we select the well-known SemanticKITTI data set~\cite{behley2019semantickitti}. Semantic labels $\mathcal{Y}_t$ for training are obtained from the Cylinder3D-multiscan model~\cite{cylinder3d} and the ground truth labels in~\cite{behley2019semantickitti} are used to \textcolor{black}{generate a ground truth world model $\mathcal{G}$}. As evaluation data for multi-scan dynamic \textcolor{black}{semantic} mapping with free space \textcolor{black}{labels} is not available, we generate it ourselves by keeping the semantically-labeled point cloud intact; but adding free space samples onto it. For evaluation of both \textcolor{black}{map }accuracy and \textcolor{black}{map} completeness, we create a point set \textcolor{black}{$\mathcal{D}_t^{\text{free}}$ containing only free space labels,} by sampling free space every 1.5m from the sensor origin to each point in the point cloud. \textcolor{black}{$\mathcal{D}_t^{\text{free}}$} is then downsampled by a voxel-grid filter and added to the ground truth semantically labeled point cloud $\mathcal{D}_t$ to generate~$\mathcal{G}$.

\textcolor{black}{For map accuracy,} to compute the ``closest semantic neighbor'' $g_m$ discussed in Section~\ref{sec:eval}, we consider the semantics of all points in $\mathcal{G}$ that fall within each visible voxel $m \in \mathcal{M}_v$. The semantic category that has the \emph{most} points in $m$ is chosen to be the ground truth semantic category \textcolor{black}{of $g_m$}. Occupied space \textcolor{black}{samples ($\mathcal{D}_t$) are given priority} over free space samples, i.e., \textcolor{black}{$g_m$ is considered to be ``free'' only if $m$ exclusively contains free space samples.}

We show results of this approach on the entire SemanticKITTI data set in Table~\ref{tab:single_view_comp} and demonstrate how we improve map accuracy over S-BKI with D-BKI.  The average IoU over each scan is computed for each sequence and aggregated per class. The IoUs of a particular class are highlighted if there is a >0.01 difference between the methods. One can see that S-BKI and D-BKI perform similarly for 20 of the static classes. However, for all the 6 dynamic classes, D-BKI consistently outperforms S-BKI by a significant margin. This result \textcolor{black}{also} shows the importance of \textcolor{black}{free space consideration for evaluating dynamic maps} \textcolor{black}{as the artifacts introduced by S-BKI seen in Fig.~\ref{fig:se1_1_qualitative} and Fig.~\ref{fig:seq_4_qualitative} would not be evaluated if we restrict our evaluation methodology only to occupied space.}

For \textcolor{black}{map} completeness, \textcolor{black}{the ``closest semantic neighbor'' $m_g$ for each $g$ in the ground truth $\mathcal{G}$ is the voxel in $\mathcal{M}$ that $g$ falls within.} To compare with other dynamic semantic mapping methods, we picked four sequences that are representative of the challenges faced while driving in dynamic environments --- highways (sequence 01) \textcolor{black}{at high speed}, countrysides (sequence 03) and cities (sequence 06) \textcolor{black}{at normal} speed, and residential areas in city (sequence 10) \textcolor{black}{at} slow speed. On this subset, we showcase in Table~\ref{tab:single_view_comp} how D-BKI and Kochanov et al.~\cite{kochanov2016scene} perform in the \textcolor{black}{map }completeness metric. We average the Jaccard scores (i.e., mean IoU) across each scan in these four sequences for both methods at the same resolution of 0.3 \textcolor{black}{m}. D-BKI performs better or similarly to Kochanov et al.~\cite{kochanov2016scene} in dynamic classes. As the map resolution is low, D-BKI's mIoU drops slightly for pedestrians (smaller objects) but remains high for larger dynamic objects. The mIoU is observed to be significantly higher in static classes than \cite{kochanov2016scene}. % To be revisited
%write individual issues
\begin{table*}[t]
% \footnotesize
\centering
\caption{\textcolor{black}{Quantitative results on Dynamic-BKI and Cylinder3D \cite{cylinder3d} for evaluation of the} \textbf{Semantic Segmentation Sub-Task} on the SemanticKITTI data set sequence 00-10 (Training) and 11-21 (Testing) ~\cite{behley2019semantickitti} for 26 semantic classes. \textcolor{black}{The collected data is evaluated with the performance metric \textbf{mean IoU (mIoU)}.}}
\label{table:semantickitti}
\resizebox{\textwidth}{!}{
\begin{tabular}{llllcccccccccccccccccccccccc}
% \toprule
% \multicolumn{3}{c|}{Method}
{\bf Seq.} & 
\multicolumn{1}{l}{\bf Method}&
\cellcolor{scarColor}\rotatebox{90}{\color{white}Car} &
\cellcolor{sbicycleColor}\rotatebox{90}{\color{white}Bicycle} &
\cellcolor{smotorcycleColor}\rotatebox{90}{\color{white}Motorcycle} &
\cellcolor{struckColor}\rotatebox{90}{\color{white}Truck} & 
\cellcolor{sothervehicleColor}\rotatebox{90}{\color{white}Other Vehicle} & 
\cellcolor{spersonColor}\rotatebox{90}{\color{white}Person} &
\cellcolor{sbicyclistColor}\rotatebox{90}{\color{white}Bicyclist} &
\cellcolor{smotorcyclistColor}\rotatebox{90}{\color{white}Motorcyclist} &
\cellcolor{sroadColor}\rotatebox{90}{\color{white}Road} &
\cellcolor{sparkingColor}\rotatebox{90}{\color{white}Parking} &
\cellcolor{ssidewalkColor}\rotatebox{90}{\color{white}Sidewalk} &
\cellcolor{sothergroundColor}\rotatebox{90}{\color{white}Other Ground} &
\cellcolor{sbuildingColor}\rotatebox{90}{\color{white}Building} &
\cellcolor{sfenceColor}\rotatebox{90}{\color{white}Fence} &
\cellcolor{svegetationColor}\rotatebox{90}{\color{white}Vegetation} &
\cellcolor{strunkColor}\rotatebox{90}{\color{white}Trunk} &
\cellcolor{sterrainColor}\rotatebox{90}{\color{white}Terrain} &
\cellcolor{spoleColor}\rotatebox{90}{\color{white}Pole} &
\cellcolor{strafficsignColor}\rotatebox{90}{\color{white}Traffic Sign} &
\cellcolor{scarColor}\rotatebox{90}{\color{white} Car-Moving} &
\cellcolor{sbicyclistColor}\rotatebox{90}{\color{white} Bicyclist-Moving} &
\cellcolor{spersonColor}\rotatebox{90}{\color{white} Person-Moving} &
\cellcolor{smotorcyclistColor}\rotatebox{90}{\color{white} Motorcylist-Moving} &

\cellcolor{sothervehicleColor}\rotatebox{90}{\color{white} Other Vehicle-Moving} &
\cellcolor{struckColor}\rotatebox{90}{\color{white} Truck-Moving} &

\rotatebox{90}{\bf Average}\\ \hline 

\vspace{-2mm} \\
\multirow{2}{*}{Training}
& Cylinder3D & 0.950 & 0.604 & 0.824 & 0.927 & 0.820 & 0.629 & n/a & n/a & 0.954 & 0.744 & 0.863 & 0.423 & 0.895 & 0.776 & 0.893 & 0.714 & 0.792 & 0.751 & 0.812 & 0.918 & 0.917 & 0.683 & 0.663 & 0.912 & 0.496 & 0.781 \\
%& S-BKI & 0.954 & 0.681 & 0.875 & 0.940 & 0.865 & 0.714 & n/a & n/a & 0.959 & 0.788 & 0.871 & 0.451 & 0.912 & 0.803 & 0.895 & 0.723 & 0.803 & 0.754 & 0.823 & 0.929 & 0.906 & 0.772 & 0.613 & 0.793 & 0.585 & 0.800 \\ note: is this necessary anymore?
& D-BKI (Ours) & 0.954 & 0.648 & 0.883 & 0.953 & 0.865 & 0.690 & n/a & n/a & 0.958 & 0.772 & 0.867 & 0.453 & 0.913 & 0.790 & 0.896 & 0.716 & 0.794 & 0.753 & 0.827 & 0.913 & 0.913 & 0.770 & 0.612 & 0.788 & 0.578 & 0.796 \\
\bottomrule
\multirow{2}{*}{Testing}
& Cylinder3D & 0.946 & 0.676 & 0.638 & 0.413 & 0.388 & 0.125 & 0.017 & 0.002 & 0.907 & 0.65 & 0.745 & 0.323 & 0.926 & 0.66 & 0.858 & 0.72 & 0.689 & 0.631 & 0.614 & 0.749 & 0.683 & 0.657 & 0.119 & 0.001 & 0.0 & 0.525 \\
& D-BKI (Ours) & 0.946 & 0.593 & 0.411 & 0.495 & 0.461 & 0.273 & 0.0 & 0.0 & 0.907 & 0.663 & 0.748 & 0.26 & 0.906 & 0.656 & 0.857 & 0.727 & 0.711 & 0.637 & 0.694 & 0.756 & 0.64 & 0.656 & 0.329 & 0.221 & 0.012 & 0.542 \\
\bottomrule
\end{tabular}
}
% \squeezeup
\label{tab:semseg_data}
\end{table*}
\noindent \par \textbf{Multi-view data set}: {For the CARLA data set, we picked five scenarios often encountered in a dynamic urban environment --- (i) a static car in the presence of moving pedestrians, (ii) a car having to stop in the presence of a jaywalking individual, (iii) a car driving at fast and (iv) slow speeds in dense traffic conditions and lastly, (v) a car driving in light traffic conditions. This data is acquired over an 1800 scan sequence and each of these scenarios is 100 scans long, but presents different challenges. The semantic segmentation labels input into the map are obtained from the simulator.}

We compare the \textcolor{black}{map accuracy between S-BKI and D-BKI in Table~\ref{tab:carla_acc}}. Note that in accuracy, each \emph{visible} voxel ($m \in \mathcal{M}_v$) is queried against the ground truth model $\mathcal{G}$. Since S-BKI and D-BKI share the same map representation, the maps share the same origin and have overlapping voxels. 

Precision is an indicator of how many predictions made by the map match the ground truth, \textcolor{black}{and is calculated per semantic class $k$ as the number of voxels $m$ correctly labeled $k$, divided by the total number of voxels $m$ labeled $k$. Therefore, precision will be lower for dynamic classes if residual traces are not removed during map propagation.} For example, the trails seen in Fig.~\ref{fig:qualitative_kitti} \textcolor{black}{for S-BKI lead to a low precision for dynamic classes. This pattern may be seen in Table \ref{tab:carla_acc}, where D-BKI has improved precision on the vehicle and pedestrian classes.}

\textcolor{black}{Recall is another useful metric for evaluating maps in dynamic environments. In contrast to precision, recall is calculated as the proportion of ground truth measurements $g_m$ with semantic label $k$ that were correctly identified. The difference in recall between static and dynamic mapping is most evident in the free class. If traces from dynamic objects are not removed, free space voxels will be incorrectly labeled occupied, and thus recall for the free category will be lower. This is also evident in Table \ref{tab:carla_acc}.}

{To evaluate how much of the ground truth $G$ is modeled \textcolor{black}{correctly} by the maps, we report \textcolor{black}{map} completeness \textcolor{black}{using both the visible and occluded portions of the environment (shown in Fig.~\ref{fig:gtmap}) that correspond to $\mathcal{M}_v$ and $\mathcal{M}_o$ in the map}. Table \ref{tab:carla_comp} showcases our map performance in comparison to the scene-propagation-based dynamic semantic mapping by Kochanov et al. \cite{kochanov2016scene}. Experiments were conducted with a map resolution of 0.1 m against a higher resolution ground truth voxel map. As our mapping method inputs the Velodyne point cloud without free space sampling, our precision in Table~\ref{table:completeness} is slightly lower due to smoothing at the boundaries of occupied space. \textcolor{black}{Our results are still comparable despite using less information.} This is especially evident for the class ``traffic sign,'' as it is a smaller object. Recall for D-BKI is higher for all occupied semantic classes in the CARLA data set. 
We also evaluate the Jaccard score of the occluded portions of the map and find that D-BKI performs better than \cite{kochanov2016scene} in all semantic categories. This experiment shows D-BKI can retain the occluded parts of the map much better over a period of time.}

\subsubsection{Auxiliary Task: Semantic Segmentation}

\textcolor{black}{For the sub-task of semantic segmentation,} we evaluate our results quantitatively on the SemanticKITTI benchmark. \textcolor{black}{The ground truth semantically-annotated point clouds are available in the data set.}

Semantic \textcolor{black}{observations} $\mathcal{Y}_t$ for training are obtained from the Cylinder3D-multiscan model \cite{cylinder3d} and the data set is divided into training (sequences 00-10) and testing (sequences 11-21). For each point in a point cloud ($\mathcal{X}_t$), we compute the per-class mean IoU using the Jaccard index against the ground truth labels provided in the SemanticKITTI data set. \textcolor{black}{Table \ref{tab:quantitative_parameters} details the parameters used to run the experiments at a map resolution of 0.1}. %\sout{Our evaluation method is limited to the occupied voxels in the map and we do not query unoccupied voxels.  Lastly, we also compare dynamic D-BKI mapping with the static mapping framework (S-BKI) to see if we have equivalent performance in the occupied voxels. For a fair comparison to S-BKI, we do not sample free space as S-BKI would become overconfident about free space over multiple scans and fail to map occupied regions accurately. In the absence of free space sampling, we remove FORC from D-BKI and set the prior on $\alpha_0^{\text{free}}$ slightly higher than the other classes. Table \ref{tab:all_data} shows the performance of our Dynamic-BKI (D-BKI) mapping on the SemanticKITTI data set.}
%write about experimental parameters in S-BKI
Our training results in Table \ref{tab:semseg_data} show that \textcolor{black}{D-BKI} mapping improves upon Cylinder3D in nearly every category. The results also showcase that spatio-temporal smoothing is beneficial for segmentation accuracy, a valuable insight for future research directions in this area. %\sout{While this demonstrates that spatio-temporal smoothing is beneficialfor segmentation accuracy, it does not showcase the full capabilities of the map, as a "full scan" only captures occupied space. The D-BKI map improves over static-BKI mapping by removing tails from dynamic objects, as demonstrated qualitatively in Section~\ref{sec:kittiqual}.}

\subsection{Discussion}
% Lessons Learned
We showed that a simple auto-regressive transition model enables dynamic scene propagation and rectifies the pitfalls of the static world assumption in the Semantic-BKI mapping algorithm - either by reducing traces in the map or by preventing overconfidence in free space. Metrics proposed to evaluate dynamic mapping quantitatively can aid in providing more perspective on the appearance of the global metric-semantic map rather than the \textcolor{black}{local} view. This can be helpful in checking whether unoccupied space is erroneously classified as occupied or vice versa.  The work can be applied to any sensor data that can be represented in the form of an XYZ point cloud. Given acquiring scene flow data for a full point cloud is more challenging than acquiring it for camera data, we anticipate that the performance is transferable to other 3D sensors.

%Important limitations
Setting the map to build at finer resolutions achieves significantly better performance, but \textcolor{black}{at the expense of higher computational burden and memory usage}. A garbage collection process when a sequence runs too long for dynamic-mapping application may be useful. Having \textcolor{black}{adaptive} kernel lengths according to object size (e.g. for vehicular and human classes) may improve results.
%\todo{Insert new discussion}
Future work includes investigating methods to compress and streamline data acquisition, demonstrating results on data sets in unstructured environments, and investigating memory-based alternatives to the autoregressive model.
% Scene flow
% Resolution and cost trade-off, limitation
% AR model is simple, could be improved
% S-KITTI not the best metric, limited by data sets
\section{Conclusion}
\label{sec:conclusion}
We developed a dynamic mapping algorithm based on Bayesian kernel inference that models the motion of dynamic objects using scene flow. Our map may be built from any 3D sensor and uses deep neural networks to obtain semantic labels and scene flow from raw point cloud data. For the evaluation of dynamic semantic maps we build, we utilize a quantitative metric that assesses semantics and unoccupied space in a unified manner. We demonstrated the efficacy of the mapping system on simulated data, a single-view real data set and multi-view synthetic data set. In particular, the proposed method, D-BKI, can reconstruct a dynamic scene more precisely than its static counterpart while maintaining high recall. D-BKI can also perform on par with state-of-the-art methods in dynamic semantic mapping in the presence of noisy as well as more accurate labels.

Future work includes investigating methods to compress and streamline data acquisition, working on data sets in unstructured environments and investigating memory-based alternatives to the autoregressive model. Exploiting the uncertainty estimated by D-BKI in scene understanding and robot navigation tasks~\cite{9813568} is also an attractive future work direction.
%, where at the time of writing this paper we achieved second in the multi-scan leaderboard.
% Future work includes investigating deep-learning or probabilistic alternatives to the auto-regressive temporal transition model, and combining networks to minimize latency while maintaining high performance. 

% Dynamic mapping can help in identifying moving objects that corrupt pose estimates in SLAM and if accurately identified, in improving ego-motion estimates and to extrapolate areas in the map that were otherwise occluded by motion. Furthermore, including semantics in the pipeline can assist with realizing a higher-level representation of objects that can be further exploited to deal with dynamic objects.

\appendices
\section{Additional Qualitative Results}

\begin{figure*}[ht]
    \centering
    \subfloat{
        \includegraphics[width=0.5\textwidth,trim={0 0 0 0 0},clip]{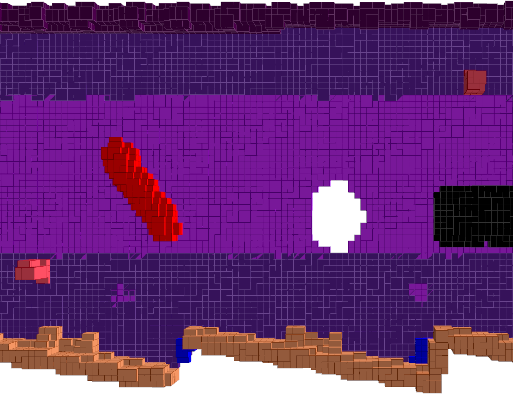}
        %\label{fig:jw_static}
    }\vline
    \subfloat{ 
    \includegraphics[width=0.5\textwidth,trim={0 0 0 0},clip]{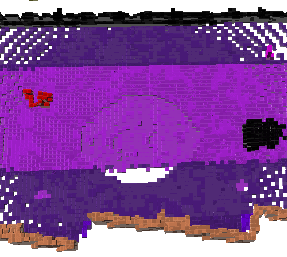}
        %\label{fig:jw_bl}
    }\\
    \subfloat{ 
    \includegraphics[width=0.5\textwidth,trim={0 0 0 0},clip]{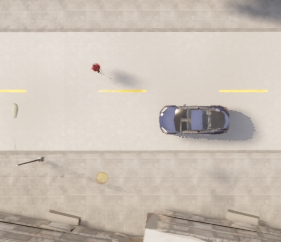}
        %\label{fig:jw_bev}
    }\vline
    \subfloat{
        \includegraphics[width=0.5\textwidth,trim={0 0 0 0},clip]{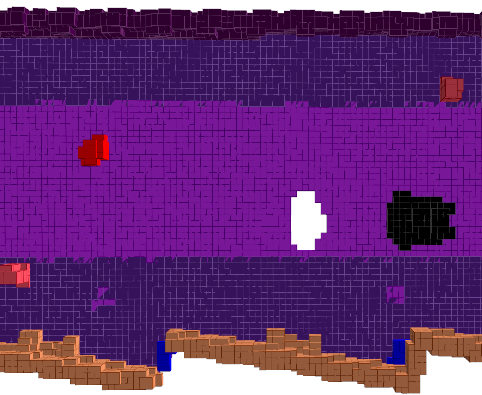}
        %\label{fig:jw_dynamic}
    }
    
    \caption{Qualitative results on S-BKI (top left), Kochanov et. al.  (top-right) and D-BKI (bottom-right) of the CARLA dataset. In this scene, there is a jaywalking pedestrian (bottom-left) and the ego-car (in dark-blue) stops for them. In all the map images, the red blob is the jaywalking pedestrian and the black rectangle is the car. One can see that D-BKI "fills" in the general area of the pedestrian's existence (more than Kochanov et al) and does not leave trails of the walking pedestrians as with S-BKI.} 
    \label{fig:qualitative_jaywalk}
\end{figure*}

\begin{figure*}[ht]
    \centering
    \subfloat{
        \includegraphics[width=0.5\textwidth,trim={0 0 0 0},clip]{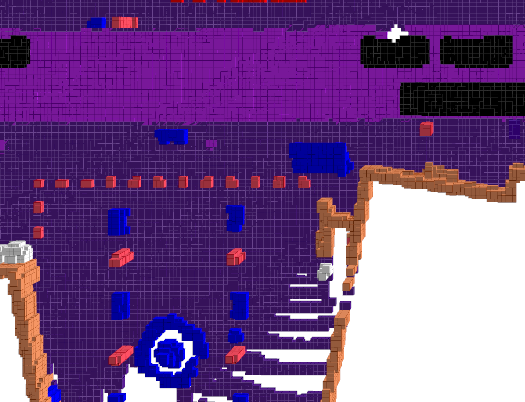}
        %\label{fig:hm_static}
    }\vline
    \subfloat{ 
    \includegraphics[width=0.5\textwidth,trim={0 0 20 0},clip]{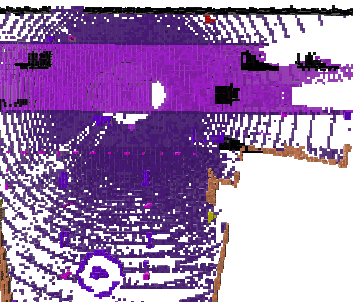}
        %\label{fig:hm_bl}
    }\\
    \subfloat{ 
    \includegraphics[width=0.5\textwidth,trim={0 0 0 0},clip]{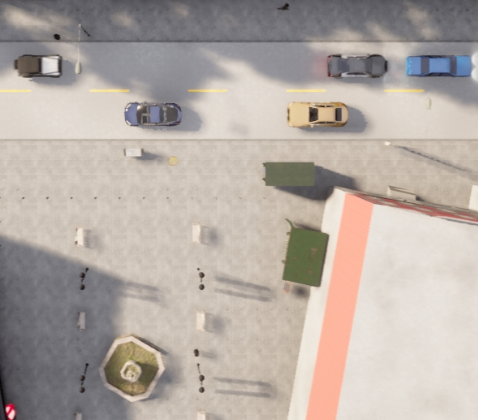}
        %\label{fig:hm_bev}
    }\vline
    \subfloat{
        \includegraphics[width=0.5\textwidth,trim={0 0 0 0},clip]{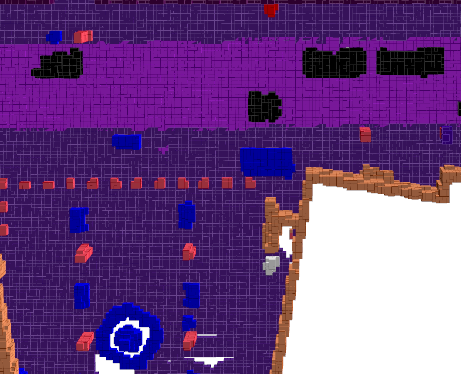}
        %\label{fig:hm_dynamic}
    }
    
    \caption{Qualitative results on S-BKI (top left), Kochanov et. al.~\cite{kochanov2016scene} (top-right) and D-BKI (bottom-right) of the CARLA dataset. This scene (bottom-left) showcases our ego-car (in dark-blue) encountering four other cars in heavier traffic. D-BKI tracks the other cars with minimum trails. It can be seen that more ``gaps'' are filled on the vehicles (in black) by D-BKI. } 
    \label{fig:qualitative_heavy_moving}
\end{figure*}

This section provides additional qualitative results shown in Fig.~\ref{fig:qualitative_jaywalk} and \ref{fig:qualitative_heavy_moving}.

{%\small
% \balance
\bibliographystyle{IEEEtran}
\bibliography{bib/strings-full,bib/ieee-full,bib/refs}
}

\begin{IEEEbiography}[{\includegraphics[width=1in,height=1.25in,clip,keepaspectratio]{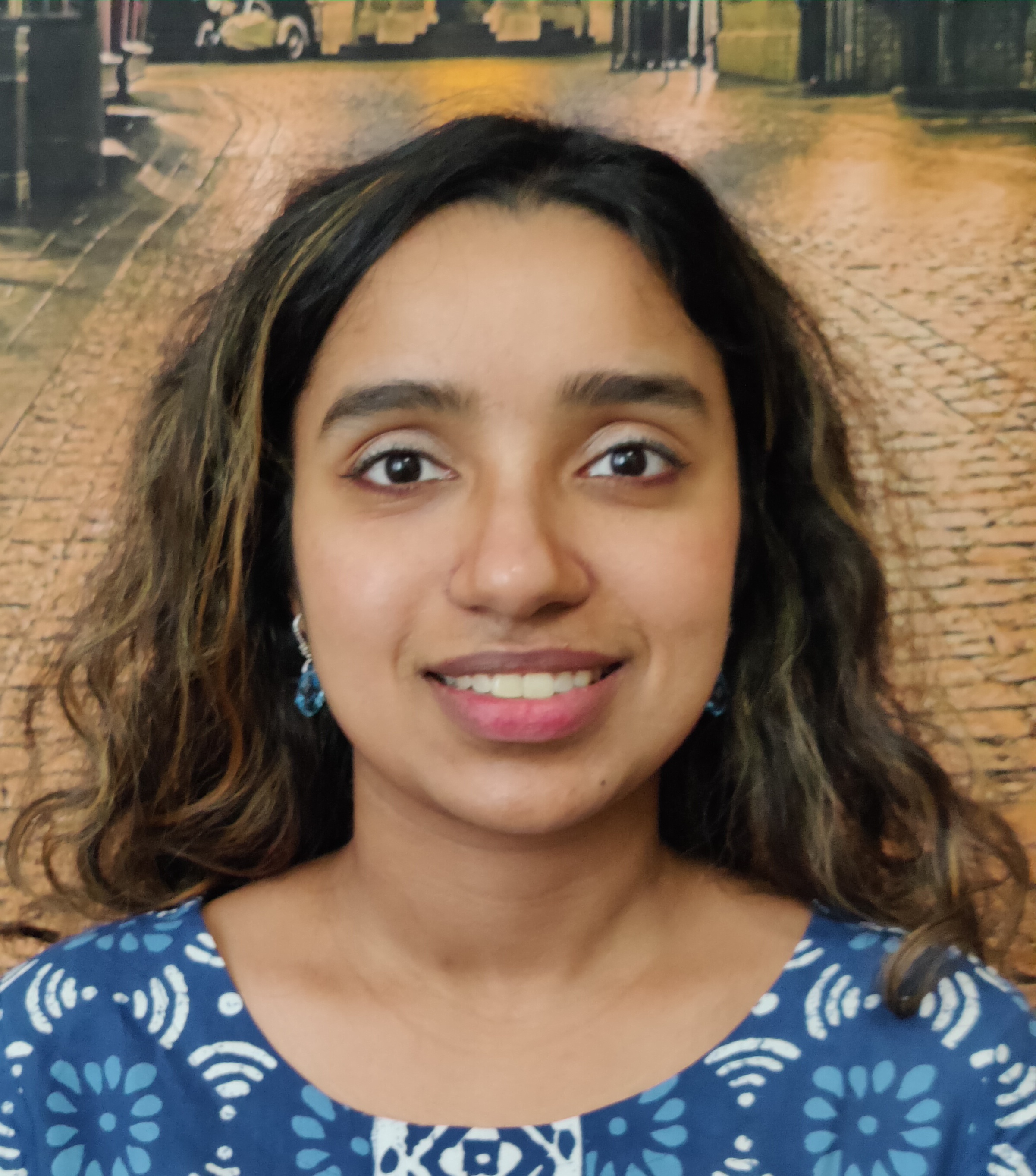}}]{Aishwarya Unnikrishnan} received a B.~Tech. degree in electrical \& electronics engineering from Delhi Technological University, Delhi, India, in 2016 and a dual-M.S. degree in Robotics \& Electrical \& Computer Engineering from the University of Michigan (UM), Ann Arbor, MI, USA in 2021. She is currently a Research Engineer at University of Michigan. Her current research interests include on-road perception and mapping.
\end{IEEEbiography}

\begin{IEEEbiography}[{\includegraphics[width=1in,height=1.25in,clip,keepaspectratio]{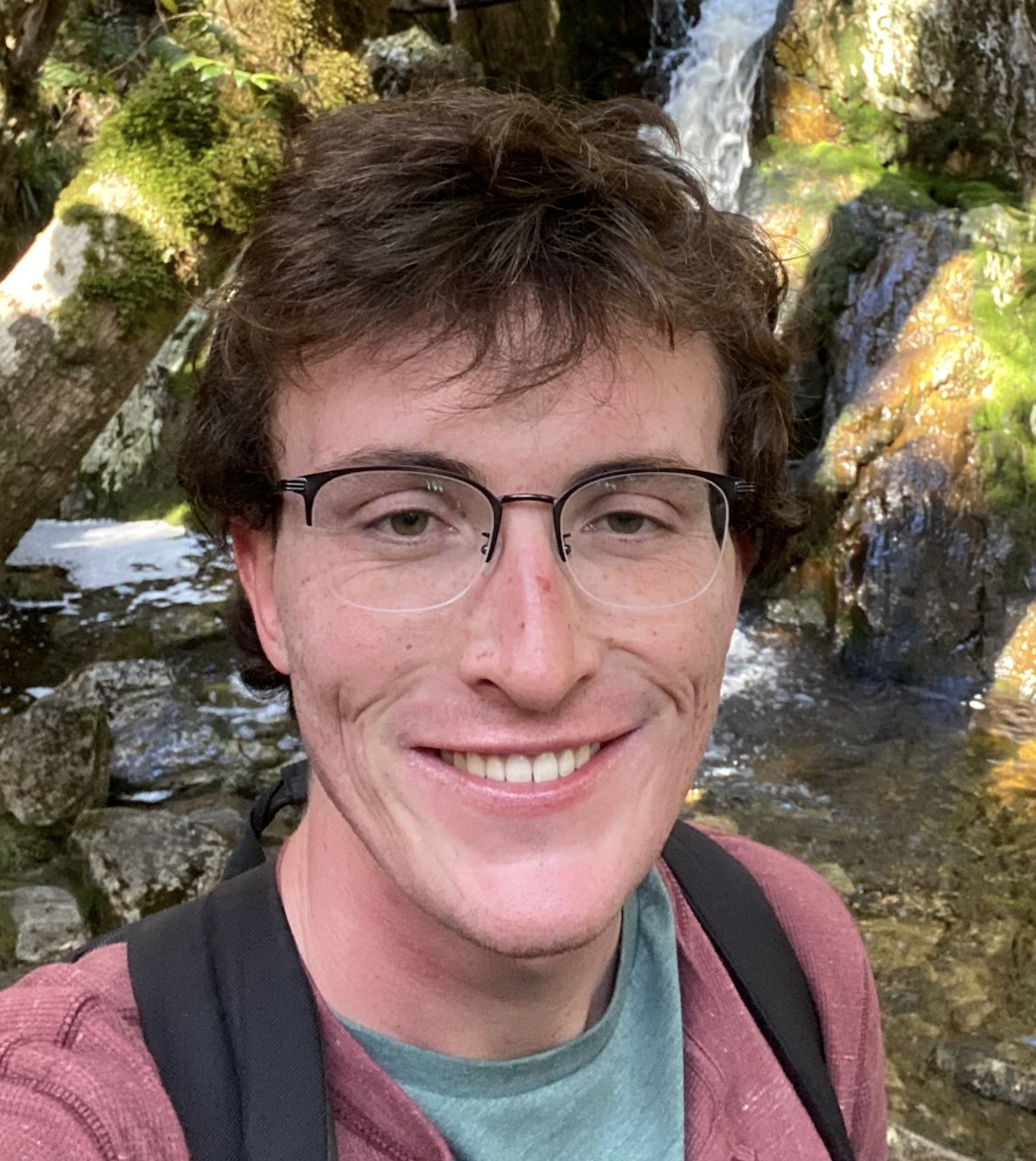}}]{Joey Wilson} received the B.S. degree in computer engineering from California Polytechnic State University San Luis Obispo (Cal Poly), CA, USA, in 2019. He is currently a Ph.D. candidate in the University of Michigan Robotics Institute, Ann Arbor, MI, USA. His current research interests include scene understanding in dynamic environments for autonomous systems.
\end{IEEEbiography}

\begin{IEEEbiography}[{\includegraphics[width=1in,height=1.25in,clip,keepaspectratio]{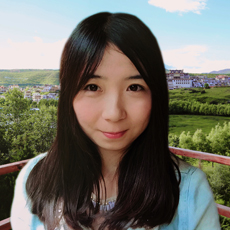}}]{Lu Gan} received the M.S. and Ph.D. degrees in Robotics from the University of Michigan (UM), Ann Arbor, MI, USA, in 2021 and 2022. She is currently a postdoctoral scholar at the Graduate Aerospace Laboratories of the California Institute of Technology (GALCIT), Pasadena, CA, USA. Her current research interests include computer vision, perception and navigation for autonomous systems.
\end{IEEEbiography}

\begin{IEEEbiography}[{\includegraphics[width=1in,height=1.25in,clip,keepaspectratio]{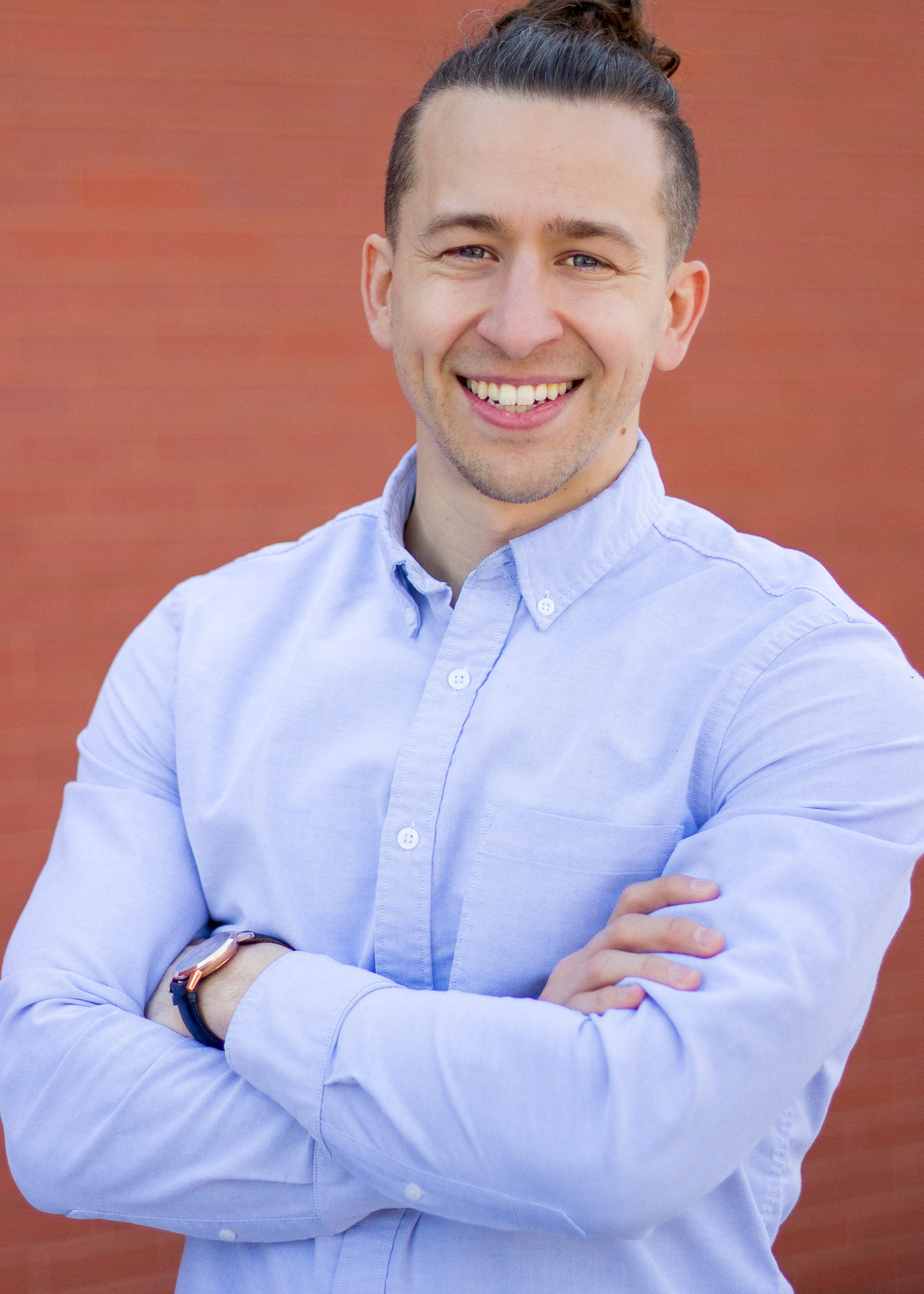}}]{Andrew Capodieci} is the Robotics Integration Group Lead at Neya Systems and has been with Neya for 9 years. During this time, Andrew has managed and performed applied research in all areas of the robotics stack including kinodynamically-feasible path planning in congested spaces, traversability estimation in off-road terrain, and object detection and classification. As Integration Group Lead, Andrew is focused on transitioning Neya's state-of-the-art autonomy research into fieldable, robust autonomy capabilities that deliver value to the warfighter. Andrew has led numerous multi-million-dollar programs including Neya's work on GVSC's Combat Vehicle Robotics program, and programs to develop autonomous construction vehicles for the commercial sector.
\end{IEEEbiography}

\begin{IEEEbiography}[{\includegraphics[width=1in,height=1.25in,clip,keepaspectratio]{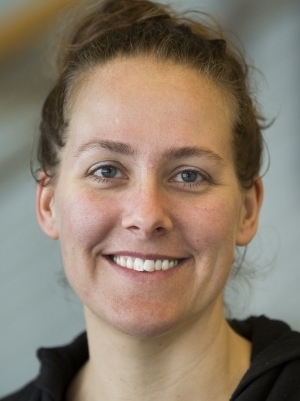}}]{Kira Barton} received the Ph.D. degree in Mechanical Engineering from the University of Illinois at Urbana-Champaign, USA, in 2010. She is currently an Associate Professor at the Robotics Institute and Department of Mechanical Engineering, University of Michigan, Ann Arbor, MI, USA. Her research interests lie in control theory and applications including high precision motion control, iterative learning control, and control for autonomous vehicles.
\end{IEEEbiography}

\begin{IEEEbiography}[{\includegraphics[width=1in,height=1.25in,clip,keepaspectratio]{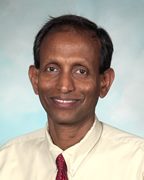}}]{Paramsothy~Jayakumar} is a Senior Technical Expert in Analytics at U.S. Army DEVCOM Ground Vehicle Systems Center (GVSC). Prior to joining GVSC, he worked at Ford Motor Company and BAE Systems. Dr. Jayakumar is a Fellow of the Society of Automotive Engineers, and the American Society of Mechanical Engineers. He is also an Associate Editor of the ASME Journal of Autonomous Vehicles and Systems, and Editorial Board Member of the International Journal of Vehicle Performance and the Journal of Terramechanics. Dr. Jayakumar has received the DoD Laboratory Scientist of the Quarter Award, NATO Applied Vehicle Technology Panel Excellence Awards, SAE Arch T. Colwell Cooperative Engineering Medal, SAE James M. Crawford Technical Standards Board Outstanding Achievement Award, BAE Systems Chairman’s Award, and NDIA GVSETS Best Paper Awards. He has published over 200 papers in peer-reviewed literature. Dr. Jayakumar received his M.S. and Ph.D. from Caltech, and B.Sc. Eng. (Hons, First Class) from the University of Peradeniya, Sri Lanka.
\end{IEEEbiography}

\begin{IEEEbiography}[{\includegraphics[width=1in,height=1.25in,clip,keepaspectratio]{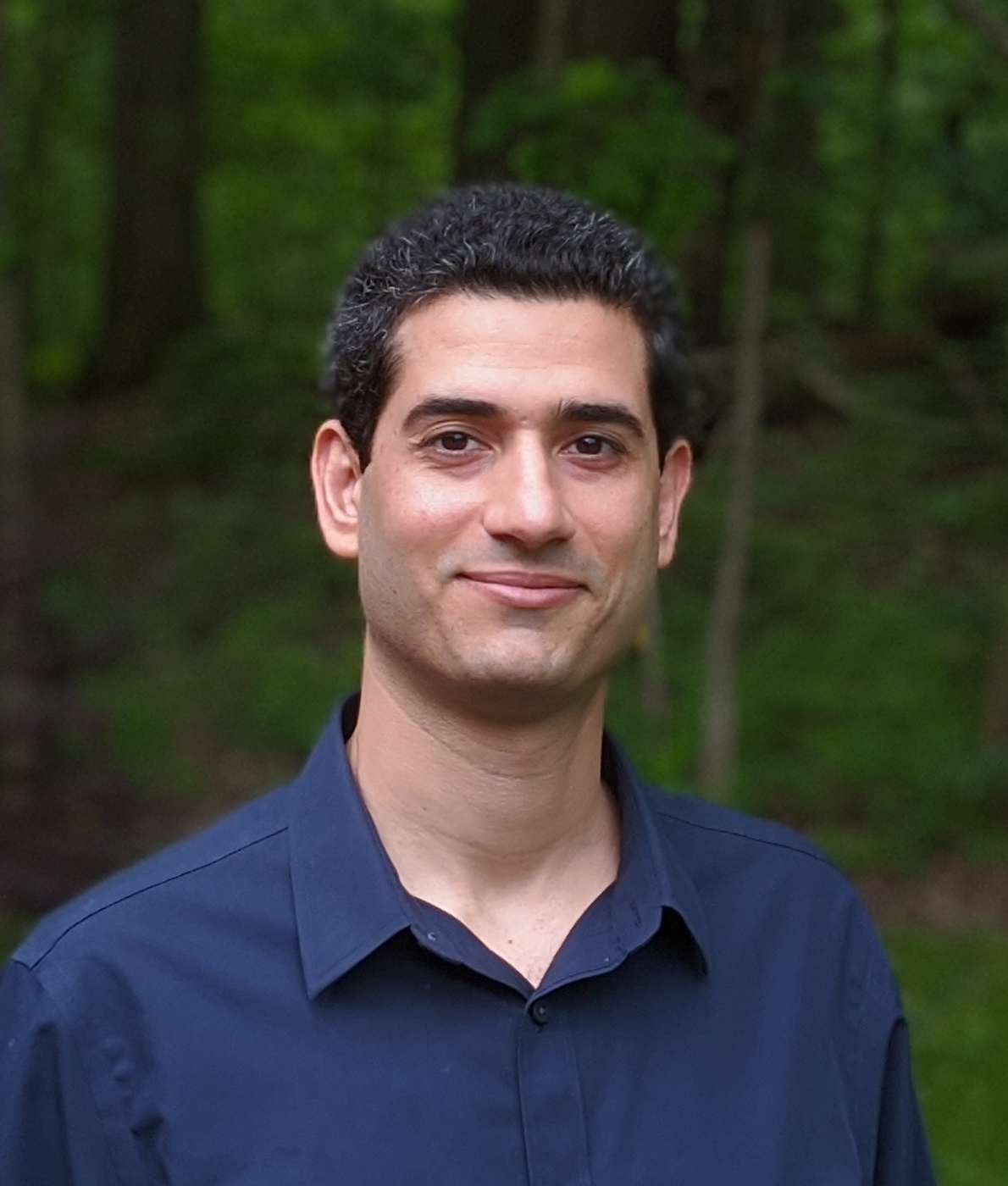}}]{Maani Ghaffari} received the Ph.D. degree from the Centre for Autonomous Systems (CAS), University of Technology Sydney, NSW, Australia, in 2017. He is currently an Assistant Professor at the Robotics Institute and Department of Naval Architecture and Marine Engineering, University of Michigan, Ann Arbor, MI, USA. He recently established the Computational Autonomy and Robotics Laboratory. His research interests lie in the theory and applications of robotics and autonomous systems.
\end{IEEEbiography}

\EOD

\end{document}